\documentclass[lettersize,journal]{IEEEtran}
\PassOptionsToPackage{dvipsnames,svgnames,table}{xcolor}
\usepackage{amsmath,amsfonts}
\usepackage{array}
\usepackage[caption=false,font=normalsize,labelfont=sf,textfont=sf]{subfig}
\usepackage{textcomp}
\usepackage{booktabs} 
\usepackage{stfloats}
\usepackage{url}
\usepackage{verbatim}
\usepackage{graphicx}
\usepackage{multirow}
\usepackage{svg}
\usepackage{array}
\usepackage{amssymb}
\usepackage{algorithm}
\usepackage{algorithmic}
\usepackage{hyperref}
\hypersetup{hypertex=true, colorlinks=true, linkcolor=red,anchorcolor=blue,citecolor=green}
\usepackage{adjustbox}
\usepackage{multirow}
\def\BibTeX{{\rm B\kern-.05em{\sc i\kern-.025em b}\kern-.08em
    T\kern-.1667em\lower.7ex\hbox{E}\kern-.125emX}}
\usepackage{balance}
\begin{document}
\title{Rein{++}: Efficient Generalization and\\ Adaptation for Semantic Segmentation\\ with Vision Foundation Models}
\author{
        Zhixiang Wei,
        Xiaoxiao Ma,                
        Ruishen Yan, 
        Tao Tu,        
        Huaian Chen$^\dagger$,
        Jinjin Zheng,\\
        Yi Jin$^\dagger$,~\IEEEmembership{Member,~IEEE},
        Enhong Chen,~\IEEEmembership{Fellow,~IEEE}
        \thanks{Zhixinag Wei, Xiaoxiao Ma, Ruishen Yan, Tao Tu,  Yi Jin,  Huaian Chen, Jinjin Zheng, and Enhong Chen are with the University of Science and Technology of China, China;
                Correspondence e-mails: zhixiangwei@mail.ustc.edu.cn, xiao\_xiao@mail.ustc.edu.cn, ranruishen@mail.ustc.edu.cn, tutao9036@mail.ustc.edu.cn, jjzheng@ustc.edu.cn, anchen@mail.ustc.edu.cn, jinyi08@ustc.edu.cn, and cheneh@ustc.edu.cn.                
                 $\dagger$: Corresponding author.}}

\markboth{Journal of \LaTeX\ Class Files,~Vol.~18, No.~9, September~2020}%
{How to Use the IEEEtran \LaTeX \ Templates}

\maketitle

\begin{abstract}
Vision Foundation Models(VFMs) have achieved remarkable success in various computer vision tasks. However, their application to semantic segmentation is hindered by two significant challenges: (1) the disparity in data scale, as segmentation datasets are typically much smaller than those used for VFM pre-training, and (2) domain distribution shifts, where real-world segmentation scenarios are diverse and often underrepresented during pre-training. To overcome these limitations, we present \textbf{Rein++}, an efficient VFM-based segmentation framework that demonstrates superior generalization from limited data and enables effective adaptation to diverse unlabeled scenarios. Specifically, Rein++ comprises a domain generalization solution Rein-G and a domain adaptation solution Rein-A. Rein-G introduces a set of trainable, instance-aware tokens that effectively refine the VFM's features for the segmentation task. This parameter-efficient approach fine-tunes less than 1\% of the backbone's parameters, enabling robust generalization. Building on the Rein-G, Rein-A performs unsupervised domain adaptation at both the instance and logit levels to mitigate domain shifts. In addition, it incorporates a semantic transfer module that leverages the class-agnostic capabilities of the segment anything model to enhance boundary details in the target domain. The integrated Rein++ pipeline first learns a generalizable model on a source domain (e.g., daytime scenes) and subsequently adapts it to diverse target domains (e.g., nighttime scenes) without any target labels. Comprehensive experiments demonstrate that Rein++ significantly outperforms state-of-the-art methods with efficient training, underscoring its roles an efficient, generalizable, and adaptive segmentation solution for VFMs, even for large models with billions of parameters. The code is available at \url{https://github.com/wloves/Rein}.
\end{abstract}

\IEEEdisplaynontitleabstractindextext
\IEEEpeerreviewmaketitle
\begin{figure*}
        \centering                
        \includegraphics[width=\linewidth]{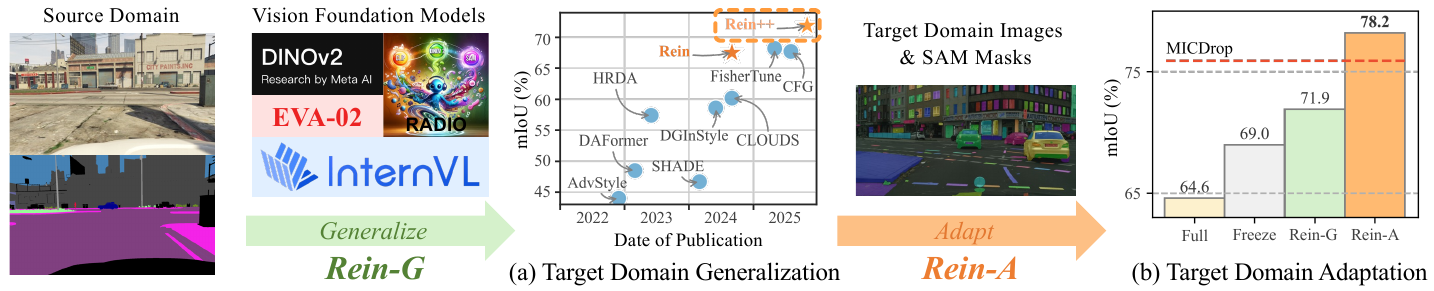}
        \caption{This paper proposes an effective training framework for both domain generalization and adaptation. We demonstrate our approach using an experiment with GTAV as source domain and Cityscapes as target domain. (a) Leveraging powerful VFMs and efficient Rein, our method achieves SoTA domain generalization performance with 71.9\% mIoU on Cityscapes validation set. (b) By incorporating target domain images, SAM masks, and upgrading Rein to domain-adaptive Rein++ for VFM adaptation, we surpass our generalization results, achieving 78.2\% mIoU for SoTA domain adaptation performance.}
        \label{fig:teaser}
\end{figure*}
\begin{figure*}
        \centering                
        \includesvg[width=\linewidth]{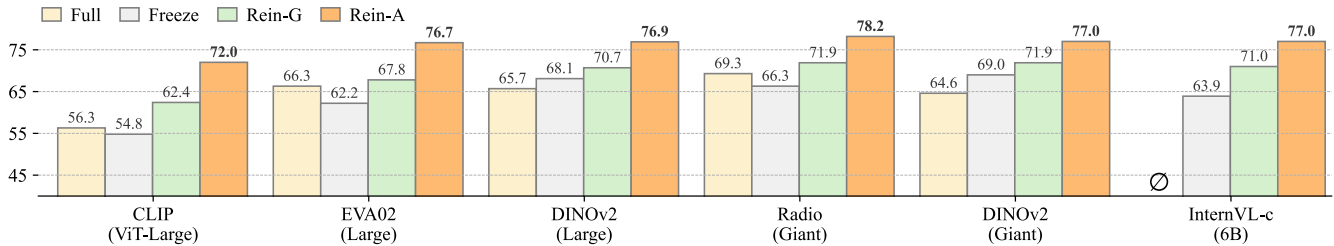}
        \caption{Using GTAV$\rightarrow$Cityscapes as a demonstration case for both generalization and adaptation, we applied Rein++ to multiple mainstream VFM families and models of varying scales. Rein++ consistently enhances generalization performance, surpassing both full finetuning and frozen backbone finetuning approaches. Simultaneously, the adaptation process significantly improves these VFMs' target domain performance. Notably, full finetuning of InternVL (6B) proves exceptionally challenging for semantic segmentation due to its massive parameter size, hence its performance data is unavailable. By contrast, Rein++ requires training only \textbf{0.4\%} of its parameters, enabling execution with approximately 24GB GPU memory while delivering substantially improved performance over frozen variants. This demonstrates Rein++'s advantage for training large models.}
        \label{fig:compare_bar_vfm}
\end{figure*}

\section{Introduction}\label{sec:introduction}

\IEEEPARstart{T}{he} emergence of large-scale Vision Foundation Models (VFMs), exemplified by models like CLIP~\cite{CLIP}, DINOv2~\cite{Dinov2}, and InternVL-c~\cite{internvl_c}, has marked a paradigm shift in computer vision, demonstrating robust generalization across numerous tasks. However, their application to semantic segmentation presents two significant challenges. 
First, the disparity in dataset scale: Semantic segmentation datasets, such as Cityscapes~\cite{cityscapes} and GTAV~\cite{gtav}, are orders of magnitude smaller than the web-scale datasets (\textit{e.g.}, LVD-142M~\cite{Dinov2}) used for VFM pre-training. Naively fine-tuning VFMs, which have a vast number of trainable parameters (ranging from approximately 300 million to 6 billion), on these smaller datasets can result in limited generalizability~\cite{scaling}.
Second, the {domain distribution shift}: Real-world segmentation scenarios are highly diverse, encompassing challenging conditions such as rain, fog, nighttime, and snow, which are underrepresented in the pre-training datasets of VFMs~\cite{samadapter}. Furthermore, annotating data for these diverse conditions is significantly more challenging than for a uniform and well-defined setting. Consequently, adapting VFMs to multiple domains without labeled data becomes a difficult and significant task.
\ifCLASSOPTIONcaptionsoff
        \newpage
\fi

For traditional networks such as VGG~\cite{vggnet} and ResNet~\cite{resnet}, these challenges have been effectively addressed. The issue of dataset scale disparity is tackled by Domain Generalization Semantic Segmentation (DGSS) methods~\cite{dg_surver,advstyle,PASTA,gtrltr,SAN-SAW,dg:tang2020selfnorm,ibn}, which train models on source domain datasets and enhance their generalization ability across unseen domains. Meanwhile, the domain distribution shift is addressed by unsupervised Domain Adaptive Semantic Segmentation (DASS) methods~\cite{tranheden2021dacs,daformer-pami,mic,micdrop,coda}, which leverage images from the target domain to improve performance under domain shift. However, these strategies have predominantly utilized CNN backbones~\cite{resnet,mobilenet} or compact ViT~\cite{xie2021segformer}, and have not fully exploited the potential of billion-parameter VFMs. To address this gap, we propose an \textbf{efficient} VFM-based segmentation framework that demonstrates superior \textbf{generalization} capabilities when fine-tuned on limited datasets, while also enabling effective \textbf{adaptation} to diverse, unlabeled scenarios.

\textbf{For Efficiency:}
To utilize VFMs efficiently in limited datasets, we opt to fine-tune VFMs with a reduced set of trainable parameters, \textit{i.e.}, parameter efficient fine-tuning (PEFT) methods. A straightforward approach involves freezing the VFMs and fine-tuning only the decoder head. Our experiments, as shown in Table~\ref{tab:inversitigate}, reveal surprising results: even with a frozen backbone, VFMs outperform most existing DG methods in terms of generalizability. However, since VFMs are typically pre-trained on tasks unrelated to segmentation, they lack task-specific knowledge for segmentation contexts such as urban scenes. Despite the efficiency, backbone freezing thus fails to fully exploit VFMs' segmentation potential.

\textbf{For Generalization:}
To enhance generalization while maintaining efficiency, we propose \textbf{Rein-G}, a robust PEFT method specifically designed for segmentation. Rein-G utilizes a minimal set of trainable parameters to leverage VFMs for superior generalization. At its core, Rein-G incorporates a set of learnable tokens, each directly associated with different instances. These tokens interact with VFM features through a dot-product operation to generate an attention-like similarity map. This map enables Rein-G to precisely refine and propagate the feature maps from each layer to the next within the backbone, significantly boosting the performance of VFMs in segmentation scenarios. Through deliberate design and token-level refinement, Rein-G achieves superior generalization compared to existing PEFT methods~\cite{lora,vpt,adapter}.

\textbf{For Adaptation:}
While Rein-G demonstrates strong generalizability, it lacks the mechanism to adapt to unlabeled domains. To overcome this limitation, we propose \textbf{Rein-A}, which extends Rein-G by leveraging its core strength: the effective encapsulation of source domain knowledge within a compact set of parameters. This enables efficient adaptation by fine-tuning only these essential components rather than the entire parameter set. Rein-A introduces three key innovations:
1) {Mask classification adaptation segmentation}: Diverging from conventional per-pixel classification~\cite{daformer-pami,mic,tranheden2021dacs,pipa,wang2021domain,ddb}, Rein-A aligns with modern VFM segmentation architectures~\cite{mask2former,maskformer,hgformer,segvit} by formulating adaptation through mask-based objectives.
2) {Dual-level supervision}: Rein-A employs complementary loss functions at both the semantic logit level and instance mask level, providing more precise gradient signals than single-level approaches and resulting in stabler convergence. 
3) Rein-A integrates SAM2~\cite{SAM2} with a semantic transfer model, transferring category-agnostic segmentation expertise to enhance boundary precision in the target domain. These key components make Rein-A an effective adaptation framework for billion-parameter VFMs.

The combination of Rein-G and Rein-A forms \textbf{Rein++}, which enables VFMs to be fine-tuned on labeled datasets, such as synthetic datasets, and effectively adapted to diverse unlabeled datasets, such as real-world datasets. A comprehensive investigation across five VFM families (CLIP, EVA02, DINOv2, Radio, and InternVL-C) and three model scales (large, giant, and 6B) demonstrates that Rein++ consistently outperforms both frozen-backbone and full fine-tuning approaches in terms of generalization ability, as illustrated in Fig.~\ref{fig:compare_bar_vfm}. Furthermore, the adaptation capabilities of Rein++ enhance segmentation performance even in the absence of labels. Rein++ offers two key advantages: (1) \textbf{Memory Efficiency}: For large-scale models such as InternVL-C (6B parameters), adaptation typically requires over 80GB of GPU memory (batch size=1), which exceeds the typical memory capacity of a single GPU. Rein++ reduces this requirement to 31GB; (2) \textbf{Storage Efficiency}: Adapting to multiple scenes only necessitates storing lightweight token sets, rather than duplicating the entire backbone. These advantages establish Rein++ as an efficient architecture for harnessing VFMs in both Domain Adaptation and Generalization settings. In summary, the main contributions of this paper are as follows:
\begin{itemize} 
    \item We first assess and harness various Vision Foundation Models (VFMs) in the context of Domain Generalization Semantic Segmentation (DGSS). Extensive experiments highlight the exceptional generalizability of VFMs and establish strong baselines for DGSS. Using proposed efficient method, we achieve Domain Adaptive Semantic Segmentation with billion-parameter VFMs for the first time.
    \item We present a robust fine-tuning method, termed \textbf{Rein-G}. With a set of learnable tokens, Rein-G refines features within each backbone layer. As a result, Rein-G enhances the capability of VFMs in DGSS tasks with fewer trainable parameters.
    \item Building on Rein-G, we propose \textbf{Rein-A}, which adapts VFMs at both the logit and instance levels. Additionally, through alignment with SAM, Rein-A effectively adapts VFMs for various segmentation tasks and scenarios.
    \item The combination of Rein-G and Rein-A forms \textbf{Rein++}, which enables efficient generalization and adaptation for semantic segmentation. Comprehensive experiments across diverse settings demonstrate that Rein++ efficiently fine-tunes VFMs for generalizable and adaptive segmentation, consistently outperforming previous DA and DG segmentation methods.
\end{itemize}

This work extends our preliminary framework Rein \cite{Rein} presented at CVPR 2024. While Rein demonstrated strong generalization capabilities, {it lacked a mechanism to leverage readily available unlabeled target domain images, limiting its full potential}. \textbf{The core contribution of this journal version is the introduction of adaptation capability}, resulting in the unified {Rein++} framework that enables domain adaptation (DA) to unlabeled target domains while maintaining domain generalization (DG) strengths. Additionally, we introduce three complementary enhancements to further boost generalization performance: (1) {Larger VFMs}: Support for Giant and 6 Billion parameter models; (2) {More VFMs}: Implementation on Radio~\cite{radio} and InternVL-C~\cite{internvl_c}; (3) {Design Improvements}: Multi-head mechanism and GeLU activation for better performance without extra parameters. {Overall, the introduction of adaptation capabilities and enhancements in generalization make Rein++ a more powerful and comprehensive framework compared to the conference version.}
\begin{table*}[htbp]
\caption{Performance benchmarking of \textbf{multiple VFMs and previous DGSS methods} under the \textit{GTAV$\rightarrow$Citys. +BDD. +Map. }\textbf{generalization} setting. Without specialized design,frozen VFMs demonstrate stronger performance.}
\label{tab:inversitigate}
\centering
\belowrulesep=0pt
\aboverulesep=0pt
\resizebox{\textwidth}{!}{
\renewcommand\arraystretch{1.1}
\setlength\tabcolsep{2.0pt}{
\begin{tabular}{l|cccccc|cccccc}
\toprule
&\multicolumn{6}{c}{Existing DGSS methods}&\multicolumn{6}{|c}{Frozen backbone of VFMs}\\
\midrule
Methods&GTR\cite{gtrltr}&AdvStyle\cite{advstyle}&TLDR\cite{TLDR}&HRDA\cite{daformer-pami}&BlineNet\cite{blindnet}&SCSD\cite{scsd}&CLIP-L\cite{CLIP}&EVA02-L\cite{EVA02}&DINOv2-l~\cite{Dinov2}&Radio-G~\cite{internvl_c}&DINOv2-G~\cite{Dinov2}&InternVL-6B~\cite{internvl_c}\\
Publications&TIP21&NIPS22&ICCV23&TPAMI23&CVPR24&AAAI25&ICML21&arXiv23&arXiv23&CVPR24&arXiv23&CVPR24\\
\midrule
mIoU(Citys.) &43.7&43.4&47.6&57.4&45.7&51.7&54.8\cellcolor[HTML]{F3F3F3}&62.2\cellcolor[HTML]{F3F3F3}&68.1\cellcolor[HTML]{F3F3F3}&66.3\cellcolor[HTML]{F3F3F3}&\textbf{69.0}\cellcolor[HTML]{F3F3F3}&{63.9}\cellcolor[HTML]{F3F3F3}\\
mIoU(BDD.)   &39.6&40.3&44.9&49.1&41.3&44.7&45.3\cellcolor[HTML]{F3F3F3}&51.3\cellcolor[HTML]{F3F3F3}&60.0\cellcolor[HTML]{F3F3F3}&53.5\cellcolor[HTML]{F3F3F3}&\textbf{57.6}\cellcolor[HTML]{F3F3F3}&{53.7}\cellcolor[HTML]{F3F3F3}\\
mIoU(Map.)   &39.1&42.0&48.8&61.2&47.1&57.0&58.6\cellcolor[HTML]{F3F3F3}&63.3\cellcolor[HTML]{F3F3F3}&68.9\cellcolor[HTML]{F3F3F3}&63.1\cellcolor[HTML]{F3F3F3}&\textbf{68.8}\cellcolor[HTML]{F3F3F3}&{65.6}\cellcolor[HTML]{F3F3F3}\\
\midrule
mIoU(Average)&40.8&41.9&47.1&55.9&44.7&51.1&52.9\cellcolor[HTML]{F3F3F3}&58.9\cellcolor[HTML]{F3F3F3}&65.5\cellcolor[HTML]{F3F3F3}&61.0\cellcolor[HTML]{F3F3F3}&\textbf{65.1}\cellcolor[HTML]{F3F3F3}&{61.1}\cellcolor[HTML]{F3F3F3}\\
\bottomrule
\end{tabular}
}}

\end{table*}
\begin{figure*}
        \includegraphics[width=\linewidth]{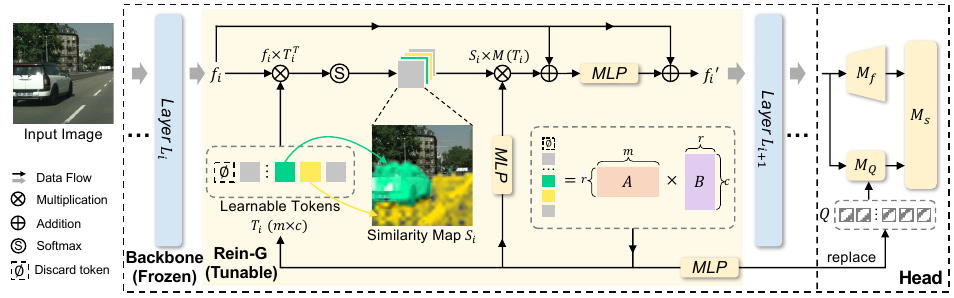}        
        \caption{An overview of proposed Rein-G. Rein-G primarily consists of a collection of low-rank learnable tokens, denoted as $T=\{T_1,T_2,\ldots,T_N\}$. These tokens establish direct connections to distinct instances, facilitating instance-level feature refinement. This mechanism results in the generation of an enhancement feature map $f'_i=f_i+Rein(f_i)$ for each layer within backbone. All parameters of MLPs are layer-shared to reduce the number of parameters. $M_f$, $M_Q$, and $M_S$ are features module, queries module, and segmentation module, respectively.}
        \label{fig:framework_rein}
\end{figure*}
\section{Related Works}
\label{related_work}
\subsection{Domain Generalized Semantic Segmentation}
Domain Generalized Semantic Segmentation (DGSS) focuses on enhancing model generalizability. This field involves training models on a set of source domains to enhance their performance on distinct and unseen target domains. Various approaches~\cite{hgformer,SPC,MoDify,OCR,SAN-SAW,kamann2020increasing,clouds} have been proposed to address this issue, with representative methods including splitting the learned features into domain-invariant and domain-specific components~\cite{dg:xu2022dirl,dg:tang2020selfnorm}, or employing meta-learning to train more robust models~\cite{PintheMem}. A standard scenario in DGSS is generalizing from one urban-scene dataset to another, for instance, from the synthetic GTAV~\cite{gtav} dataset to the real-world Cityscapes~\cite{cityscapes}. In this classic setting, certain techniques~\cite{choi2021robustnet,ibn,dg:switchable} have achieved notable performance through learning feature normalization/whitening schemes, while others~\cite{wildnet} have improved segmentation results through feature-level style transfer and the introduction of additional data. Additionally, strong data augmentation~\cite{advstyle,gtrltr,PASTA,fan2023towards} often simply and effectively enhances model robustness. However, most of the previous DGSS methods generally utilize outdated backbones like ResNet~\cite{resnet}, VGGNet~\cite{vggnet}, MobileNetV2~\cite{mobilenet}, and ShuffleNetV2~\cite{shufflenet}, thereby leaving the efficacy of stronger Vision Foundation Models (VFMs) in DGSS relatively unexplored.

Our preliminary work, Rein~\cite{Rein}, was one of the earliest concurrent works exploring VFMs for DGSS, alongside VLTSeg~\cite{vltseg} and CLOUDS~\cite{clouds}. Subsequently, numerous VFM-based DGSS methods have emerged, including CDG~\cite{cdg}, FADA~\cite{fada}, SET~\cite{SET}, FisherTune~\cite{fishertune}, and SoMA~\cite{soma}. Notably, \textbf{all of these methods utilize Rein's codebase or follow its implementation details.}

\subsection{Domain Adaptive Semantic Segmentation}
Unsupervised domain adaptation (UDA) aims to transfer knowledge from a label-rich domain to an unlabeled domain in the presence of domain shift. Inspired by the theoretical analysis of Ben-David et al. \cite{ben2006analysis}, the existing UDA methods usually either match a well-defined moment-based distribution discrepancy (e.g., maximum mean discrepancy \cite{tzeng2014deep}, joint maximum mean discrepancy \cite{long2017deep}, and margin disparity discrepancy \cite{zhang2019bridging} , etc.) or leverage the adversarial learning on various distributions (e.g., feature distribution \cite{ganin2016domain,chen2022reusing,wei2021toalign}, joint distribution \cite{NEURIPS2018_ab88b157,tang2020discriminative,zhang2019domain}, and gradient distribution of features \cite{gao2021gradient,du2021cross}) to learn domain-invariant features. In contrast to these direct adaptation approaches, some recent works have achieved impressive results when dealing with significant domain discrepancies by constructing domain bridging at the image \cite{na2021fixbi,bousmalis2017unsupervised,gong2019dlow,wu2020dual,Englert_2024_CVPR,Englert_2025_CVPR}, or feature level \cite{cui2020gradually,Li_2021_CVPR}. After the proposal of SAM~\cite{SAM}, recent works have paid attention to how to harness SAM for domain adaptive segmentation. GoodSAM introduces a teacher assistant head to bridge the capacity gap between SAM and the student during training~\cite{goodsam}. Inspired by existing works~\cite{goodsam,samadapter}, this paper introduces SAM as a supervision on the target domain.

\subsection{Vision Foundation Models}
The concept of a Foundation Model, initially introduced by ~\cite{vfms} in the field of Natural Language Processing (NLP), is defined as ``the base models trained on large-scale data in a self-supervised or semi-supervised manner that can be adapted for several other downstream tasks". While models like the ViT~\cite{vit} and Swin Transformer~\cite{swin} have demonstrated excellent performance, the quest for a Vision Foundation Model (VFM) similar to their NLP counterparts is ongoing. This pursuit has yielded significant advancements with the advent of models such as CLIP~\cite{CLIP}, which learn high-quality visual representations by exploring contrastive learning with large-scale image text pairs; MAE~\cite{MAE}, utilizing a masked image modeling for learning latent image representations; SAM~\cite{SAM}, which develops a promptable model and pre-train it on a broad dataset for segmentation task; EVA02~\cite{EVA,EVA02}, which integrates Masked Image Modeling pre-training with CLIP's vision features; and DINOv2~\cite{Dinov2}, which is pretrained on extensive, curated datasets without explicit supervision. These VFMs have shown remarkable performance in downstream applications. Yet, a dedicated investigation into their performance in the specific context of DGSS and DASS tasks remains unexplored.

\subsection{Parameter-Efficient Fine-tuning}
In the realm of NLP, parameter-efficient fine-tuning (PEFT) has achieved notable success by freezing most parameters of VFMs and fine-tuning a select few. Various approaches have been developed, such as BitFit~\cite{Bitfit}, which adjusts only the model's bias terms; Prompt-tuning~\cite{prompt_tuning}, introducing soft prompts to adapt frozen language models; Adapter-tuning~\cite{adapter}, adding lightweight modules to each transformer layer; and notably, LoRA~\cite{lora}, which injects trainable rank decomposition matrices into transformer layers, yielding significant influence.

The application of PEFT methods is also expanding into the field of computer vision~\cite{fahes2023simple,kumar2022fine}, with notable examples such as Visual Prompt Tuning (VPT)~\cite{vpt}, which prepends prompts into the input sequence of transformer layers; AdaptFormer~\cite{adaptformer}, replacing the MLP block in the transformer encoder with an AdaptMLP; LP-FT~\cite{kumar2022fine} finds that fine-tuning can achieve worse accuracy than linear probing out-of-distribution; and Prompt-ICM~\cite{feng2023prompt}, applying large-scale pre-trained models to the task of image coding for machines. 
Contrasting with these methods, we aim to refine feature maps for each instance within an image, thereby achieving superior performance in the realm of DGSS.

\begin{figure*}[htbp]
        \centering
        \includegraphics[width=\linewidth]{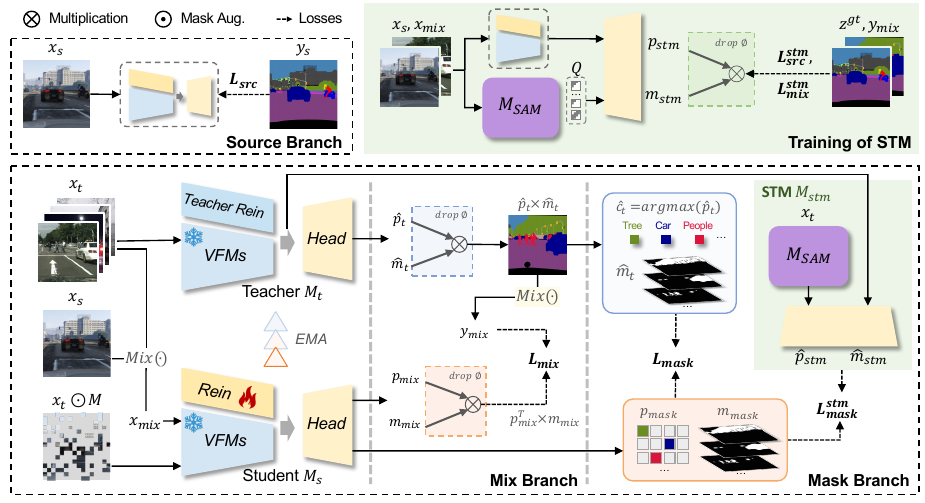}        
        \caption{Overview of the Rein-A framework. Rein-A is a mask classification-based semantic segmentation approach that generates mutually aligned masks and class logits. These instance-level masks are subsequently aggregated to form comprehensive semantic segmentation maps. To enable stable training in unlabeled target domains, Rein-A simultaneously incorporates: 1) a mix branch trained on cross-domain mixed images, and 2) a mask branch trained on masked target images. Furthermore, Rein-A introduces a Semantic Transfer Module (STM) that transfers semantic-agnostic mask knowledge from Segment Anything Model (SAM) to the segmentation model, yielding refined mask boundaries. Detailed pipeline implementation is provided in Algorithm~\ref{alg:MCAS}.}
        \label{fig:framework_mcas}
\end{figure*}

\section{Introducing the Rein++ Framework}
This work focuses on VFMs for semantic segmentation, particularly targeting large-scale VFMs. To effectively fine-tune such parameter-intensive VFMs on limited datasets, we propose \textbf{Rein-G}, which achieves significant performance improvements while training only 1\% of the VFM's parameters, demonstrating exceptional generalization capabilities. Building upon Rein-G, we introduce \textbf{Rein-A} – an unsupervised domain adaptation method tailored for VFMs – to address diverse real-world scenarios lacking labeled data. By confining knowledge learned from the source domain to a compact set of parameters, Rein-G enables Rein-A to perform stable and efficient unsupervised fine-tuning, further boosting the VFM's performance in target domains. The synergistic integration of Rein-G and Rein-A gives rise to \textbf{Rein++}, a comprehensive framework that: (1) obtains a highly generalizable semantic segmentation model on a common labeled domain (e.g., daytime scenes), and (2) unsupervisedly enhances model performance on a significantly different unlabeled domain (e.g., nighttime scenes). The core mechanisms and implementation details of Rein-G are elaborated in Section~\ref{sec:rein-g}, while Section~\ref{sec:rein-a} presents the framework and implementation of Rein-A.

\section{Rein-G: Generalized Fine tuning}
\label{sec:rein-g}
To utilize VFMs efficiently in limited datasets, we opt to fine-tune VFMs with a reduced set of trainable parameters. A straightforward thought might involve a smaller decode head; however, this method merely acts as a passive receiver of feature maps from the backbone, lacking the flexibility to effectively adapt a frozen backbone for generating task-specific or scene-specific features. In contrast, we propose to embed a mechanism, named ``Rein-G", between the layers within the backbone. Rein-G actively refines and forwards the feature maps from each layer to the subsequent one. This approach allows us to more effectively utilize the powerful capabilities of VFMs, much like using rein to control a horse.

Given a pre-trained VFM with parameters $\Phi_{M}$, consisting of a sequence of layers $L_1,L_2,\ldots,L_N$, a decode head $\mathcal{H}$ parameterized by $\theta_{h}$, and the Rein-G strategy with parameters $\theta_{R}$, the optimization objective can be written as:
\begin{equation}
    \label{eq:peft_optimal}
    \mathop{\arg\min}\limits_{\theta_{R},\theta_{h}}
    \sum_{i=1}^{N_d} \mathcal{L}oss(\mathcal{H}_{\theta_{h}}(\mathcal{F}_{\Phi_{M},\theta_{R}}(x_i)),y_i),
\end{equation}
where $x_i$ and $y_i$ denote the input image and its corresponding ground truth, respectively, and $N_d$ signifies the total number of samples. $\mathcal{F}_{\Phi_{M},\theta_{R}}$ represents the forward process of VFM after applying the Rein-G strategy.

\subsection{Core of Rein-G}
\label{sec:mainbody}
For simple implementation across different VFMs, we opt not to modify MLP weights at specific positions as described in the~\cite{adaptformer,lora}. Instead, our approach focuses on refining the output feature maps at each layer within the VFMs, as illustrated in Fig.~\ref{fig:framework_rein}. Precisely, for the features $f_i$ produced by the $i$-th layer $L_i$, Rein-G produces enhanced feature maps for the next layer as follows:
\begin{equation}
    \begin{aligned}
        f_{1}   & =L_{1}~(Embed(x))~~~~~~~~~~f_{1}\in \mathbb{R}^{n\times c}, \\
        f_{i+1} & =L_{i+1}(f_{i}+\Delta f_{i})~~~~~~~~~~i=1,2,\ldots,N-1,     \\
        f_{out} & =f_N+\Delta f_N,
    \end{aligned}
    \label{eq:delta}
\end{equation}
where $f'_i=f_{i}+\Delta f_{i}$ symbolizes the refined feature map, $x$ is the input image, $Embed$ denotes the patch embedding layer in VFMs, $n$ represents the number of patches, $N$ denotes the number of layers, and $c$ is the dimensionality of $f_1,f_2,\ldots,f_N$. Note that the layers $L_1,L_2,\ldots,L_N$ are kept frozen, and our focus is on training an efficient module, Rein, to generate $\Delta f_{i}$ as follows:
\begin{equation}
    \Delta f_i=Rein(f_i)~~~~~~~~ \Delta f_i\in \mathbb{R}^{n\times c},i=1,2,\dots,N.
\end{equation}

In the context of DGSS, an ideal $\Delta f_{i}$ should assist VFMs to bridge two types of gaps. The first is gap in scene between pre-training dataset and target scene, exemplified by the contrast between ImageNet~\cite{imagenet} and urban-scene images~\cite{cityscapes,gtav}. The second is task divergence between pre-training and fine-tuning, such as the differences between masked image modeling and semantic segmentation tasks.

To establish this dual bridge, Rein-G starts with a set of learnable tokens $T=\{T_i\in \mathbb{R}^{m\times c}~|~i\in\mathbb{N},1\leq i\leq N\}$, where each token sequence $T_i$ is randomly initialized, and $m$ denotes the sequence length of $T_i$. Rein-G freezes the backbone and embeds knowledge learned from the fine-tuning dataset into these tokens, thereby bridging the gap in scene relative to the pre-training dataset. Moreover, considering the essential need in semantic segmentation to discern multiple instances within a single image, Rein-G implements an attention-inspired mechanism, which enables VFMs to make tailored adjustments to the features of distinct instances, thereby aiding VFMs in adapting to the differences between semantic segmentation and pre-training tasks. Specifically, Rein-G employs a dot-product operation to generate a similarity map $S_i$, which captures the associations between feature vectors in $f_i$ and the tokens in $T$:
\begin{equation}
    \label{eq:dotproduct}
    S_i=f_i\times T_i^\text{T}~~~~~~~~ S_i\in \mathbb{R}^{n\times m},
\end{equation}
where $T_i$ represents the token sequence of the $i$-th layer, $m$ indicates the number of tokens in $T_i$. As $S$ quantitatively evaluates the relationships between various tokens and feature vectors, Rein-G can apply a softmax function to align each patch with a unique instance:
\begin{equation}
    \label{eq:softmax}
    S_i=Softmax(\frac{f_i\times T_i^\text{T}}{\sqrt{c}}).
\end{equation}

Leveraging the feature-to-token similarity map $S_i$, we can preliminarily estimates of $\Delta f_i$ using the equation:
\begin{equation}
    \label{eq:obtain_delta}
    \Delta \bar{f_i}=S_i\times [~T_i\times W_{T_i} + b_{T_i}],
\end{equation}
where $W_{T_i}$ and $b_{T_i}$ denote the weights and biases of a MLP, respectively. This MLP enables the transformation of $T_i$ across different feature spaces during the computation of $S_i$ and $\Delta \bar{f_i}$. Optionally, Rein-G can pre-calculate $T_i\times W_{T_i}+b_{T_i}$ to reduce inference time.
However, directly using the softmax-normalized $S_i$ (where the sum of each row equals one) can induce unnecessary changes when all features are already accurate. To mitigate this, we discard the first token in both $S_i$ and $T_i$. This allows the sum of each row in $S_i$ to vary between 0 and 1, reducing the risk of introducing spurious modifications.

To enhance the flexibility in feature adjustment, Rein-G utilizes a MLP composed of $W_{f_i}$ and $b_{f_i}$ to produce the final feature modifications $\Delta f_i$:
\begin{equation}
    \label{eq:finaldelta}
    \Delta f_i=(\Delta \bar{f_i}+f_i) \times W_{f_i} +b_{f_i}.
\end{equation}

Benefiting from these instance-level $\Delta f_i$ adjustments, Rein-G is capable of generating diverse modifications for various categories within a single image. The details of Rein-G will be explained in the next:

\subsection{Details of Rein-G}
\label{sec:details_of_rein}
\textbf{Linking tokens to instances.} At the core of Rein, we establish an implicit yet effective linkage between tokens and instances, which has demonstrated notable performance, as detailed in Sec.~\ref{sec:experiments}. This connection is further reinforced by utilizing object queries, a key component in DETR\cite{detr}-style decode heads~\cite{maskformer,mask2former,segvit}, as intermediaries. These queries are empirically proven to establish a direct association with instances. Specifically, we generate layer-wise queries $Q_i$ from our learnable tokens $T_i$ via linear transformation:
\begin{equation}
    \label{eq:link}
    Q_i=T_i \times W_{Q_i}+b_{Q_i}~~~~~~~~Q_i\in\mathbb{R}^{m\times c'},
\end{equation}
where $W_{Q_i}$ and $b_{Q_i}$ signify the weights and biases, respectively, and $c'$ denotes the dimension of $Q_i$. However, due to the complexity arising from the large numbers of various layers in VFMs, transforming the diverse $Q_i$ into a single query $Q$ poses computational challenges. To address this, Rein-G computes both the maximal component $Q_{max} \in \mathbb{R}^{m \times c'}$ and the average component $Q_{avg} \in \mathbb{R}^{m \times c'}$ using the following equation:
\begin{equation}
    \begin{aligned}
        Q_{max}(j,k) & =\max_{i=1,2,\ldots,N}{Q_i(j,k)},   \\
        Q_{avg}(j,k) & =\frac{1}{N}\sum_{i=1}^N{Q_i(j,k)}.
    \end{aligned}
    \label{eq:max_avg}
\end{equation}
Subsequently, $Q$ is derived as:
\begin{align}
    \label{eq:link2}
    Q & =Concat([Q_{max},Q_{avg},Q_N])\times W_Q + b_Q.
\end{align}
By mapping $T$ onto $Q$, which subsequently links to instances, Rein-G achieves enhanced performance with a marginal increase in parameters.

\textbf{Layer-shared MLP weights.} To address the redundancy of parameters in the layer-specific MLP weights, specifically $W_{T_i}$ in Eq.~(\ref{eq:obtain_delta}), $W_{f_i}$ in Eq.~(\ref{eq:finaldelta}), and $W_{Q_i}$ in Eq.~(\ref{eq:link}), which collectively contribute to a substantial trainable parameter count, we adopt a new strategy. Since the learnable $T_i$ is capable of producing distinct $\Delta f_i$ for each layer, we design the role of the MLP to primarily perform consistent linear transformations across different feature spaces for each layer within the backbone. To this end, we employ shared MLP weights across layers as outlined in the equations:
\begin{equation}
    \begin{aligned}
         & [W_{T_1},b_{T_1}]=[W_{T_2},b_{T_2}]=\ldots=[W_{T_N},b_{T_N}],                       \\
         & [W_{f_1},b_{f_1}]~=[W_{f_2},b_{f_2}]~~=\ldots=[W_{f_N},b_{f_N}],                    \\
         & [W_{Q_1},b_{Q_1}]=[W_{Q_2},b_{Q_2}]=\ldots=[W_{Q_N},b_{Q_N}].\label{eq:shareweight}
    \end{aligned}
\end{equation}

\textbf{Low-rank token sequence.} Recognizing the potential for information overlap among diverse learnable tokens, such as the high similarity between tokens representing a car's headlight and a bicycle's light, Rein-G adopts a strategy to generate a low-rank token sequence $T$ as follows:
\begin{equation}
    \label{eq:lowrank}
    Ti=A_i\times B_i, ~~~~~~~A\in \mathbb{R}^{m\times r}, B\in \mathbb{R}^{r\times c},
\end{equation}
where $c$ denotes the dimension of $T_i$, $m$ is the length of sequence $T_i$, and $r$ represents the rank, with $r\ll c$. Here, matrices $A$ and $B$ are constructed as low-rank matrices. To reduce inference time, Rein-G can precompute and store $T$. By implementing this low-rank token sequence approach, Rein-G significantly reduces the number of parameters.

\textbf{Multi-head and Non-linearity.}
Following classical attention architectures~\cite{attention}, we enhance the \textit{Rein} module through two synergistic modifications without parameter overhead: (1) Multi-head Mechanism: Implementing a parallel multi-head mechanism akin to~\cite{attention}, we decompose feature interactions into multiple subspaces while preserving the original parameter count.  (2) Nonlinear Activation Function: Introducing a GELU activation within the MLP layer augments representational power. This modifies Equation~(\ref{eq:finaldelta}) as:
\begin{equation}
    \Delta f_i = W_{g} \cdot \text{GELU}\left( (\Delta \bar{f} + f_i) \cdot W_{f} + b_{f} \right) + b_{g}.
\end{equation}

Through dimensional constraints $W_{f} \in \mathbb{R}^{c \times \frac{c}{2}}$ and $W_{g} \in \mathbb{R}^{\frac{c}{2} \times c}$, we maintain the total parameters of \textit{Rein} unchanged. This dual enhancement enables richer feature interactions and nonlinear mapping while strictly preserving parameter efficiency.  
\begin{algorithm}[t!]
    \caption{The pipeline of Rein-A. For ease of understanding, the training process for the STM is outlined in \textcolor{gray}{gray} text.}
    \label{alg:MCAS}
    \begin{algorithmic}[1] 
    \REQUIRE Source image $x_s$, source label $y_s$, target image $x_t$.
    \REQUIRE Student model $\mathcal{M}_s$, teacher model $\mathcal{M}_t$, Semantic Transfer Model $\mathcal{M}_{stm}$, Segment Anything Model $\mathcal{M}_{SAM}$
    \ENSURE Adapted student network $\mathcal{M}_s$.
    \WHILE{not converged}
        \STATE \textbf{Generate Pseudo Labels}
        \STATE $(\hat{p}_t, \hat{m}_t) \gets \mathcal{M}_t(x_t)$
        \STATE {$(\hat{p}_{stm}, \hat{m}_{stm}) \gets \mathcal{M}_{stm}[x_{t}, \mathcal{M}_{SAM}(x_{t})]$ }
        \STATE $(\hat{c}_t,\hat{c}_{stm}) \gets [argmax(\hat{p}_t),argmax(\hat{p}_{stm})]$
        \STATE \textbf{Source Branch}
        \STATE $\mathcal{L}_{src} \gets \mathcal{L}_{mask-cls}(\mathcal{M}_s(x_s), y_s)$ \COMMENT{Eq. \ref{eq:maskcls}}
        \STATE \textcolor{gray}{$\mathcal{L}_{src}^{stm} \gets \mathcal{L}_{mask-cls}\{\mathcal{M}_{stm}[x_s, \mathcal{M}_{SAM}(x_s)], y_s\}$ \COMMENT{Eq. \ref{eq:stm_learn}}}
        \STATE \textbf{Mix Branch}
        \STATE $x_{mix},y_{mix} \gets Mix(x_s,y_s, x_t,\hat{p}_t^T\times\hat{m}_t)$
        \STATE $(p_{mix}, m_{mix}) \gets \mathcal{M}_s(x_{mix})$        
        \STATE $\mathcal{L}_{mix} \gets \mathcal{L}_{ce}(p_{mix}^T \times m_{mix}, y_{mix})$ \COMMENT{Eq. \ref{eq:logitloss}}
        \STATE \textcolor{gray}{$(p_{stm}, m_{stm}) \gets \mathcal{M}_{stm}[x_{mix}, \mathcal{M}_{SAM}(x_{mix})]$ }
        \STATE \textcolor{gray}{$\mathcal{L}_{mix}^{stm} \gets \mathcal{L}_{ce}(p_{stm}^T \times m_{stm}, y_{mix})$ \COMMENT{Eq. \ref{eq:stm_learn}}}
        \STATE \textbf{Mask Branch}
        \STATE $M \gets RandomMask()$
        \STATE $(p_{mask}, m_{mask}) \gets \mathcal{M}_s(x_t \odot M)$        
        \STATE $\mathcal{L}_{mask} \gets \mathcal{L}_{ce}(p_{mask},\hat{c}_t ) + \mathcal{L}_{mask}(m_{mask}, \hat{m}_t)$ \COMMENT{Eq. \ref{eq:instanceloss}}
        \STATE $\mathcal{L}_{mask}^{stm} \gets \mathcal{L}_{ce}(p_{mask}, \hat{c}_{stm}) + \mathcal{L}_{mask}(m_{mask}, \hat{m}_{stm})$ \COMMENT{Eq. \ref{eq:stm_teach}}
        \STATE \textbf{Total Loss}
        \STATE $\mathcal{L} \gets \mathcal{L}_{src} + \alpha\mathcal{L}_{mix} + \beta\mathcal{L}_{mask} + \mathcal{L}_{src}^{stm} + \alpha\mathcal{L}_{mix}^{stm} + \beta\mathcal{L}_{mask}^{stm}$
        \STATE Update $\mathcal{M}_s$ and $\mathcal{M}_{stm}$ parameters by minimizing $\mathcal{L}$
        \STATE Update $\mathcal{M}_t$ by EMA
    \ENDWHILE
    \end{algorithmic}
\end{algorithm}

\section{Rein-A: Unsupervised Domain Adaptation}
\label{sec:rein-a}

Although Rein-G exhibit superior generalization ability, it lacked a mechanism to leverage readily available unlabeled target domain images, limiting its full potention. To address it, we introduce Rein-A for unsupervised domain adaptation.

For domain adaptation semantic segmentation (DASS), we denote the source domain as ${D_s} = \{ (x_s^{(i)},y_s^{(i)})\} _{i = 1}^{{N^s}}$ with $N^s$ samples drawn from the source domain $\mathcal{S}$, where $x_s^{(i)} \in {X_s}$ is an image, $y_s^{(i)} \in {Y_s}$ is the corresponding pixel-wise one-hot label covering $K$ classes. Similarly, the unlabeled target domain set is denoted as ${D_t} = \{ x_t^{(i)}\} _{i = 1}^{{N^t}}$ with $N^t$ samples drawn from the target domain $\mathcal{T}$.


Existing DASS methods often rely on per-pixel classification, as seen in architectures like DeepLabV3+~\cite{deeplabv3p} and UperNet~\cite{upernet}. However, recent advancements in semantic segmentation, particularly with the emergence of Vision Foundation Models (VFMs)~\cite{SAM,hgformer}, have highlighted the advantages of mask classification, such as MaskFormer~\cite{maskformer}, Mask2Former~\cite{mask2former}, Mp-former~\cite{mpformer}, and SegViT~\cite{segvit}.
This motivates our exploration of adapting VFMs for domain adaptation using a mask classification strategy.
We adopt the widely used Mask2Former head as the decoding component within Rein-A. Mask2Former defines the desired output $z$ as a set of $N$ probability-mask pairs, denoted as $z=\{(p_i,m_i)\}_{i=1}^N$. It utilizes a bipartite matching-based assignment $\sigma$ to compute the mask classification loss between $z$ and the ground truth $z^{gt}=\{(c_i^{gt},m_i^{gt}|c_i^{gt}\in\{1,...,K\},m_i^{gt}\in\{0,1\}^{H\times W}\}_{i=1}^{N_{gt}}$, as shown in Eq.~\ref{eq:maskcls}:

\begin{equation}
    \label{eq:maskcls}
    \mathcal{L}_{mask-cls}=\sum_{j=1}^N[-\log{p_{\sigma(j)}}(c_j^{gt})+\textbf{1}_{c_{gt}^j\neq \varnothing }\mathcal{L}_{mask}(m_{\sigma(j)},m_j^{gt})],
\end{equation}
where $\mathcal{L}_{mask}$ represents the mask loss function, $c_j^{gt}$ denotes the semantic class label. Further implementation details can be found in~\cite{mask2former}.

Training Mask2Former for domain adaptation poses a greater challenge, particularly in the absence of target domain label supervision. Directly adapting per-pixel classification approaches, such as those proposed in~\cite{daformer-pami} and~\cite{ddb}, proves ineffective and destabilizes training. This is likely due to the inherent complexity associated with mask classification segmentation. To mitigate this issue and stabilize the adaptation process, we propose a three-pronged strategy. First, we opt to train both the Rein-G module and the Mask2Former head while freezing the backbone. This significantly reduces the number of trainable parameters, thus enhancing stability. Second, we perform fine-tuning at both the logit and instance levels to further stabilize the model's adaptation capabilities. Finally, we incorporate additional target domain knowledge by leveraging the powerful segment anything model (SAM)~\cite{SAM}. This integration provides valuable cues for the model to better align with the target domain characteristics. We elaborate on these strategy in the following section and provide the complete procedure in Algorithm~\ref{alg:MCAS} and Fig.~\ref{fig:framework_mcas}.

\subsection{Logit-Level Loss}
The first branch of Rein-A is the class-mix branch, which is optimized by a logit-level loss.
Class-mix augmentation, employed in existing DA methods like DAFormer \cite{daformer-pami} and DDB \cite{ddb}, introduces instability when used directly with the mask classification loss in Eq.~\ref{eq:maskcls}. In contrast, per-pixel cross-entropy loss exhibited robustness. We hypothesize that this discrepancy arises because class-mixed images often contain incomplete instances, leading to inconsistent supervision signals from the instance-level loss. To mitigate this, we utilize per-pixel cross-entropy loss for this branch. Specifically, following the class-mix procedure outlined in \cite{daformer-pami}, we blend source labels ($y_s$) and target teacher predictions ($\hat{p}_t \times \hat{m}_t$) to generate mixed labels ($y_{mix}$). The logit-level loss is then computed as:

\begin{equation}
    \label{eq:logitloss}
    \mathcal{L}_{mix}=\mathcal{L}_{ce}(p_{mix}^T\times m_{mix},y_{mix}),
\end{equation}
where $\mathcal{L}_{ce}$ denotes the cross-entropy loss, $p_{mix}$ and $m_{mix}$ represent the predicted class probabilities and mask, respectively, generated by the student network. The details of confidence weighting are consistent with \cite{daformer-pami}.

\subsection{Instance-Level Loss}
The second branch of Rein-A is the mask segmentation branch, which is optimized by an instance-level loss.
In contrast to class-mix augmentation, while mask augmentation disrupts instance completeness in the image domain, it maintains this completeness in the label domain. We observed that applying Eq.~\ref{eq:maskcls} to masked target images proves effective. Specifically, we randomly generate masks $M$ and calculate the loss between the teacher predictions $(\hat{p}_t,\hat{m}_t)=\mathcal{M}_t(x_t)$ and the student predictions $(p_{mask},m_{mask})=\mathcal{M}_s(x_t\odot M)$, as shown in Eq.~\ref{eq:instanceloss}:

\begin{equation}
    \label{eq:instanceloss}
    \mathcal{L}_{mask}=\mathcal{L}_{ce}(p_{mask},argmax(\hat{p}_t))+\mathcal{L}_{mask}(m_{mask},\hat{m}_t),
\end{equation}
where $\mathcal{L}_{mask}$ represents the binary mask loss~\cite{maskformer}. This formulation avoids the complex bipartite matching process. Instead, it directly calculates the loss between the student and teacher outputs based on the query order, thus maintaining instance-level supervision while enhancing the stability of the adaptation.

\subsection{Semantic Transfer Module}
\label{sec:stm}
To push the performance boundaries of our model, we incorporate the powerful segmentation priors from the Segment Anything Model (SAM)~\cite{SAM,SAM2}. 
While the SAM demonstrates remarkable segmentation performance and generalizability, it lacks semantic prediction capabilities. To bridge this gap and leverage SAM for adaptation, we introduce a novel Semantic Transfer Module (STM). The STM receives semantic features from the backbone of the teacher model and segment masks from SAM predictions, enabling it to refine the teacher’s knowledge using SAM and transfer this knowledge to the student model. The STM is trained on the source branch using ground truth labels and on the mix branch using teacher model predictions, respectively. In the mask branch, the STM’s predictions supervise the student model. The STM’s forward pipeline employs a mask classification head that utilizes backbone features as input and SAM predictions as queries. This process, detailed in the top-right corner of Fig.~\ref{fig:framework_mcas}, is formulated as:
\begin{equation}
\label{eq:ta}
\mathcal{M}_{stm}(x,m)=\mathcal{M}_{stm}(M_{Backbone}(x), m),
\end{equation}
where $m=M_{SAM}(x)$. The training process of the STM is shown in the gray part of Algorithm~\ref{alg:MCAS}, and the corresponding losses are calculated by Eq.~\ref{eq:stm_learn}:
\begin{equation}
    \begin{aligned}
        \mathcal{L}_{src}^{stm} & =\mathcal{L}_{mask-cls}\{\mathcal{M}_{stm}[x_{s},\mathcal{M}_{SAM}(x_{s})],z^{gt}\}, \\
        \mathcal{L}_{mix}^{stm} & =\mathcal{L}_{ce}\{\mathcal{M}_{stm}[x_{mix},\mathcal{M}_{SAM}(x_{mix})],y_{mix}\},  \\
    \end{aligned}
    \label{eq:stm_learn}
\end{equation}
and guide the student model with the following loss:
\begin{equation}
    \mathcal{L}_{mask}^{stm}=\mathcal{L}_{ce}(p_{mask},argmax(\hat{p}_{stm}))+\mathcal{L}_{mask}(m_{mask},\hat{m}_{stm}).
    \label{eq:stm_teach}
\end{equation}

The complete training procedure of Rein-A is shown in Algorithm~\ref{alg:MCAS}. The total loss is given in Eq.~\ref{eq:totalloss}:
\begin{equation}
    \label{eq:totalloss}
    \mathcal{L} = \mathcal{L}_{src} + \alpha\mathcal{L}_{mix} + \beta\mathcal{L}_{mask} + \mathcal{L}_{src}^{stm} + \alpha\mathcal{L}_{mix}^{stm} + \beta\mathcal{L}_{mask}^{stm}
\end{equation}
\begin{table}[htbp]
    \centering
    \definecolor{awardblue}{RGB}{0,0,147}  
    \definecolor{awardorange}{RGB}{204,0,0}
    \newcommand{\first}[1]{{\textbf{#1}}}
    \newcommand{\second}[1]{{\underline{#1}}}  
    \caption{Performance comparison with the previous DGSS methods. Top-performing results are presented in \textbf{bold}, with \underline{second best} additionally emphasized. All evaluations are conducted on the validation sets.}
    \label{tab:comparison_dg}      
    \renewcommand\arraystretch{1.1}
    \begin{tabular}{lc|ccc|>{\columncolor{gray!10}}c}
    \hline
    \multicolumn{6}{c}{\textbf{Synthetic-to-Real: \textit{GTAV $\rightarrow$ Citys.+BDD.+Map.}}}\\
    \hline
     Methods&Publication&Citys.&BDD.&Map.&Avg. \\
     \hline
     GTR-LTR~\cite{gtrltr}&TIP21&43.7&39.6&39.1&40.8\\
     AdvStyle~\cite{advstyle}&NeurIPS22&44.0&40.0&42.7&42.2\\
     DAFormer~\cite{daformer-pami}&TPAMI23&52.3&47.9&54.7&51.7\\
     HRDA~\cite{daformer-pami}&TPAMI23&57.4&49.1&61.2&55.9\\
     SHADE~\cite{SHADE}&CVPR24&46.7&43.7&45.5&45.3\\
     DGInStyle~\cite{dginstyle}&CVPR24&58.6&52.3&62.5&57.8\\
     CLOUDS~\cite{clouds}&CVPR24&60.2&57.4&67.0&61.5\\
     Rein~\cite{Rein}&CVPR24&67.5&58.8&68.3&64.9\\
     FADA~\cite{fada}&NeurIPS24&68.2&61.9&68.1&66.1\\
     TQDM~\cite{tqdm}&ECCV24&68.8&59.1&70.1&66.1\\     
     FisherTune~\cite{fishertune}&CVPR25&68.2&63.3&68.7&66.6\\
     SoMA~\cite{soma}&CVPR25&71.8&61.3&\second{71.6}&68.2\\     
     MFuser~\cite{mvfuser}&CVPR25&70.2&63.1&71.3&68.2\\

     
     \hline
     \multicolumn{2}{l|}{Rein-G + CLIP-Large~\cite{CLIP}}&62.4&51.4&63.8&59.2\\
     \multicolumn{2}{l|}{Rein-G + EVA02-Large~\cite{EVA02}}&67.8&57.2&68.6&64.5\\
     \multicolumn{2}{l|}{Rein-G + DINOv2-Large~\cite{Dinov2}}&70.7&61.1&70.8&67.5\\
     \multicolumn{2}{l|}{Rein-G + Radio-Giant~\cite{radio}}&\first{71.9}&\second{62.9}&\first{71.7}&\first{68.8}\\
     \multicolumn{2}{l|}{Rein-G + DINOv2-Giant~\cite{Dinov2}}&\first{71.9}&\first{62.3}&{71.1}&\second{68.4}\\
     \multicolumn{2}{l|}{Rein-G + InternVL-6B~\cite{internvl_c}}&71.0&59.8&69.6&66.8\\

     \hline
     \multicolumn{6}{c}{\textbf{Synthetic-to-Real: \textit{GTAV+Urbansyn+Synthia $\rightarrow$ Citys.+BDD.+Map.}}}\\
    \hline
     Methods&Publication&Citys.&BDD.&Map.&Avg. \\
     \hline
     DeepLabV3+\cite{urbansyn}&arXiv23&67.9&45.4&61.5&58.3\\  SegFormer~\cite{urbansyn}&arXiv23&70.5&51.5&67.2&63.1\\     
     Rein~\cite{Rein}&CVPR24&78.4&58.6&72.1&69.7\\
     \hline
     \multicolumn{2}{l|}{Rein-G + CLIP-Large~\cite{CLIP}}&71.0&53.1&68.5&64.2\\
     \multicolumn{2}{l|}{Rein-G + EVA02-Large~\cite{EVA02}}&75.7&59.1&72.3&69.0\\
     \multicolumn{2}{l|}{Rein-G + DINOv2-Large~\cite{Dinov2}}&\second{78.5}&61.5&\second{74.9}&71.6\\
     \multicolumn{2}{l|}{Rein-G + Radio-Giant~\cite{radio}}&\second{78.5}&\second{62.1}&\first{75.0}&\second{71.9}\\
     \multicolumn{2}{l|}{Rein-G + DINOv2-Giant~\cite{Dinov2}}&\first{79.0}&\first{63.7}&{74.7}&\first{72.5}\\
     \multicolumn{2}{l|}{Rein-G + InternVL-6B~\cite{internvl_c}}&77.0&61.0&73.3&70.4\\

     \hline
     \multicolumn{6}{c}{\textbf{Real-to-Real: \textit{ Citys$\rightarrow$ ACDC (val)+BDD.+Map.}}}\\
    \hline
     Methods&Publication&ACDC&BDD.&Map.&Avg. \\
     \hline
     HRDA~\cite{daformer-pami}&TPAMI23&59.7&58.5&68.3&62.2\\
     SHADE~\cite{SHADE}&CVPR24&-&51.0&60.7&-\\
     DGInStyle~\cite{dginstyle}&CVPR24&61.0&58.8&68.0&62.6\\
     Rein~\cite{Rein}&CVPR24&71.1&63.1&76.0&70.1\\
     TQDM~\cite{tqdm}&ECCV24&71.3&64.7&76.2&70.7\\
     FADA~\cite{fada}&NeurIPS24&71.5&65.1&75.8&70.9\\
     SCSD~\cite{scsd}&AAAI25&-&52.3&62.5&-\\
     MaskViM~\cite{maskvim}&AAAI25&-&56.4&66.6&-\\
     \hline
     \multicolumn{2}{l|}{Rein-G + CLIP-Large~\cite{CLIP}}&62.4&51.4&63.8&59.2\\
     \multicolumn{2}{l|}{Rein-G + EVA02-Large~\cite{EVA02}}&67.9&60.8&73.8&67.5\\
     \multicolumn{2}{l|}{Rein-G + DINOv2-Large~\cite{Dinov2}}&71.9&\second{65.0}&\second{76.8}&\second{71.0}\\
     \multicolumn{2}{l|}{Rein-G + Radio-Giant~\cite{radio}}&\second{72.2}&\first{66.0}&\first{76.9}&\first{71.7}\\
     \multicolumn{2}{l|}{Rein-G + DINOv2-Giant~\cite{Dinov2}}&\first{72.5}&64.6&75.8&\second{71.0}\\
     \multicolumn{2}{l|}{Rein-G + InternVL-6B~\cite{internvl_c}}&68.1&63.8&74.8&68.9\\
     \hline
    \end{tabular}
\end{table}
\begin{table*}[htbp]
\newcolumntype{a}{c}
\centering
\caption{Performance Comparison of the Proposed \textbf{Rein-G} Method and Other Methods (\textbf{DGSS} and \textbf{PEFT}). The Calculation of Trainable Parameters Includes Only the Backbone.}
\label{tab:comparison_dg2}
\renewcommand\arraystretch{1.1}
\setlength\tabcolsep{3.5pt}
\begin{tabular}{ll|c|ccca|ccca|ccca}
\hline
\multirow{2}{*}{Backbone} & \multirow{2}{*}{Methods} & \multirow{2}{*}{\begin{tabular}[c]{@{}c@{}}Trainable\\ Params\end{tabular}} & \multicolumn{4}{c|}{GTAV (source)}& \multicolumn{4}{c|}{\begin{tabular}[c]{@{}c@{}}Urbansyn + Synthia + GTAV\end{tabular}} & \multicolumn{4}{c}{Cityscapes (source)}\\
\cline{4-15}
&& & Citys& BDD& Map& Average & Citys& BDD& Map& Average & ACDC (val) & BDD& Map& Average \\
\hline

\multirow{11}{*}{\begin{tabular}[c]{@{}l@{}}DINOv2\\(Large)\\~\cite{Dinov2}\end{tabular}}& Full & 304M& 65.7 & 53.0& 66.2 & 61.7 &75.9&55.6&72.3&67.9& 68.9 & 61.3 & {76.0} & 68.7 \\
& + AdvStyle~\cite{advstyle}& 304M& 66.5 & 54.8 & 67.0 & 62.8 &76.9&58.5&{74.2}&69.9&70.8&62.2&75.3&69.5\\
& + PASTA~\cite{PASTA} & 304M& 67.2 & 55.2 & 67.1 & 63.2 &-&-&-&-&71.2&62.2&75.2&69.5\\
& Freeze & 0M& 68.1 & 60.0 & 68.9 & 65.6 &77.0&59.5&73.4&70.0& 71.5 & 63.5 & 75.4 & 70.1 \\
& + AdvStyle~\cite{advstyle}& 0M& {69.5} & {60.1} & 69.2 & {66.2} &77.0&60.1&73.6&70.2& {71.8} & 64.1 & 75.3 & 70.4 \\
& + PASTA~\cite{PASTA} & 0M& 68.3 & 59.8 & 69.0 & 65.7 &-&-&-&-& 71.4 & 64.1 & 74.9 & 70.1 \\
& + BitFit~\cite{Bitfit}& 0.27M &69.0&58.5&{69.4}&65.7&{77.6}&{59.7}&73.5&{70.3}&71.3&63.8&\textbf{76.4}&{70.5}\\
& + LoRA~\cite{lora}& 0.79M &67.3&59.5&69.2&65.3&77.1&60.0&73.2&70.1&71.6&{64.4}&75.4&{70.5}\\
& + AdaptFormer~\cite{adaptformer} & 3.17M & 68.6& 57.5&68.3 &64.8 &75.8 &58.8 &{74.2} &69.6 &69.6 &63.4 &75.9 &69.6 \\
& + VPT~\cite{vpt} & 3.69M &66.7&57.9&68.6&64.4&75.3 &59.2 &73.3 &69.3 &68.6 &60.8 &74.7 &68.0 \\
\rowcolor{gray!10}
& + Rein-G& 2.99M & \textbf{70.7} & \textbf{{61.1}} & \textbf{{70.8}} & \textbf{{67.5}} &\textbf{{78.4}}&\textbf{61.5} &\textbf{{74.9}} &\textbf{{71.6}} & \textbf{{71.9}} & \textbf{{65.0}} & {{76.0}} & \textbf{{71.0}} \\
\hline

\multirow{2}{*}{\begin{tabular}[c]{@{}l@{}}CLIP~\cite{CLIP}\\(ViT-Large)\end{tabular}}& Full& 304M & 56.3&42.7&60.0&53.0&70.2&48.4&\textbf{68.7}&62.5&63.2&52.7&\textbf{71.5}&62.5\\
&Freeze&0M&54.8& 45.3&58.6&52.9&63.8&48.3&64.1&58.7&56.0&52.8&65.3&58.0\\
\rowcolor{gray!10}
& + Rein-G&2.99M&\textbf{62.4}&\textbf{51.4}&\textbf{63.8}&\textbf{59.2}&\textbf{71.0}&\textbf{53.1}&68.5&\textbf{64.2}&\textbf{63.4}&\textbf{57.2}&71.2&\textbf{63.9}\\
\hline
\multirow{2}{*}{\begin{tabular}[c]{@{}l@{}}EVA02~\cite{EVA02}\\(Large)\end{tabular}}&Full&304M&66.3&53.8&67.4&62.5&75.1&57.0&\textbf{72.7}&68.3&\textbf{69.4}&\textbf{62.4}&\textbf{75.3}&\textbf{69.0}\\
&Freeze&0M&62.2&51.3&63.3&58.9&69.4&53.2&67.5&63.4&59.1&55.3&69.5&61.3\\
\rowcolor{gray!10}
& + Rein-G&2.99M&\textbf{67.8}&\textbf{57.2}&\textbf{68.6}&\textbf{64.5}&\textbf{75.7}	&\textbf{59.1}&72.3&\textbf{69.0}&67.9&60.8&73.8&67.5\\
\hline


\multirow{2}{*}{\begin{tabular}[c]{@{}l@{}}Radio~\cite{radio}\\(Giant)\end{tabular}}& Full & 1.14B& 69.3&59.3&70.6&66.4& 78.0&60.5&74.4&71.0 &70.9&64.4&{76.3}&70.5 \\
& Freeze & 0M&66.3&53.5&63.1&61.0&76.0&59.8&69.9&68.5&64.1&54.7&71.4&63.4\\
\rowcolor{gray!10}
& + Rein-G& 6.36M &\textbf{{71.9}}&\textbf{{62.9}}&\textbf{{71.7}}&\textbf{{68.8}}&\textbf{{78.5}}&\textbf{{62.1}}&\textbf{{75.0}}&\textbf{{71.9}}&\textbf{{72.2}}&\textbf{{66.0}}&\textbf{{76.9}}&\textbf{{71.7}}\\
\hline
\multirow{2}{*}{\begin{tabular}[c]{@{}l@{}}DINOv2~\cite{Dinov2}\\(Giant)\end{tabular}}& Full & 1.14B& 64.6 & 54.7& 66.5 & 62.0 &76.3&57.1&72.5&68.6& 69.0 & 61.3 & \textbf{{76.0}} & 68.6 \\
& Freeze & 0M&69.0&57.6&68.8&65.1&76.7&{61.7}&73.1&70.5&71.3&63.8&75.1&70.1\\
\rowcolor{gray!10}
& + Rein-G& 6.36M & \textbf{{71.9}}&\textbf{{62.2}}&\textbf{{71.1}}&\textbf{{68.4}}&\textbf{{79.0}}&\textbf{{63.7}}&\textbf{{74.7}}&\textbf{{72.5}}&\textbf{{72.5}}&\textbf{{64.6}}&75.8&\textbf{{71.0}}\\
\hline
\multirow{1}{*}{\begin{tabular}[c]{@{}l@{}}InternVL-c (\textbf{6B})~\cite{radio}\end{tabular}}&Freeze & 0M&63.9&53.7&65.6&61.1&72.1&57.2&69.4&66.3&63.4&57.9&71.7&64.3\\
\rowcolor{gray!10}
& + Rein-G& 24.0M &\textbf{{71.0}}&\textbf{59.8}&\textbf{69.6}&\textbf{66.8}&\textbf{77.0}&\textbf{61.0}&\textbf{73.3}&\textbf{70.4}&\textbf{68.1}&\textbf{63.8}&\textbf{74.8}&\textbf{68.9}\\
\hline
\end{tabular}
\end{table*}
\begin{table*}[tbp]    
    \caption{Results on \textbf{Cityscapes $\rightarrow$ ACDC (test) and Cityscapes-C (level-5)}  datasets, utilizing a batch size of 8.}
    \label{tab:comparison_dg_c2cc_c2acdc}    
    \resizebox{\textwidth}{!}{%
        \renewcommand\arraystretch{1.1}
        \setlength\tabcolsep{2.5pt}
        \begin{tabular}{l|cccc|>{\columncolor{gray!10}}c|cccc|cccc|cccc|cccc|>{\columncolor{gray!10}}c}
        \hline
        \multirow{3}{*}{Target} &
        \multicolumn{5}{c|}{ACDC\cite{acdc} (test)} &
        \multicolumn{17}{c}{Cityscapes-C\cite{michaelis2019benchmarking} (level-5)} \\
        \cline{2-23}
        &
        \multirow{2}{*}{Night} &
        \multirow{2}{*}{Snow} &
        \multirow{2}{*}{Fog} &
        \multirow{2}{*}{Rain} &
         &
        \multicolumn{4}{c|}{Blur} &
        \multicolumn{4}{c|}{Noise} &
        \multicolumn{4}{c|}{Digital} &
        \multicolumn{4}{c|}{Weather} &
         \\
        \cline{7-22}
        &
        &
        &
        &
        &
        \multirow{-2}{*}{All}&
        Motion &
        Defoc &
        Glass &
        Gauss &
        Gauss &
        Impul &
        Shot &
        Speck &
        Bright &
        Contr &
        Satur &
        JPEG &
        Snow &
        Spatt &
        Fog &
        Frost &
        \multirow{-2}{*}{Avg.}
        \\
        \hline
        HGFormer &
        52.7 &
        68.6 &
        69.9 &
        72.0 &
        67.2\cellcolor[HTML]{F3F3F3}&
        64.1 & 67.2 & 61.5 & 63.6 & \textbf{27.2} & \textbf{35.7} & \textbf{32.9} & 63.1 & 79.9 & 72.9 & 78.0 & 53.6 & 55.4 & 75.8 & 75.5 & 43.2&59.4\cellcolor[HTML]{F3F3F3}
        \\
        Rein-G &
        \textbf{70.6} &
        \textbf{79.5} &
        \textbf{76.4} &
        \textbf{78.2} &
        \textbf{77.6}\cellcolor[HTML]{F3F3F3}&
        \textbf{68.5}
        &\textbf{71.7}
        &\textbf{69.7}
        &\textbf{68.7}
        &~6.2
        &23.0
        &13.1
        &\textbf{63.7}
        &\textbf{81.5}
        &\textbf{78.9}
        &\textbf{80.6}
        &\textbf{68.8}
        &\textbf{63.8}
        &\textbf{73.6}
        &\textbf{79.5}
        &\textbf{47.9}
        &\textbf{60.0}\cellcolor[HTML]{F3F3F3}
        \\
        \hline
        \end{tabular}%
    }    
\end{table*}
\begin{figure*}[htbp]
    \centering
    \setlength{\abovecaptionskip}{0.1cm}
    \setlength{\belowcaptionskip}{0.1cm}
    \includegraphics[width=\linewidth]{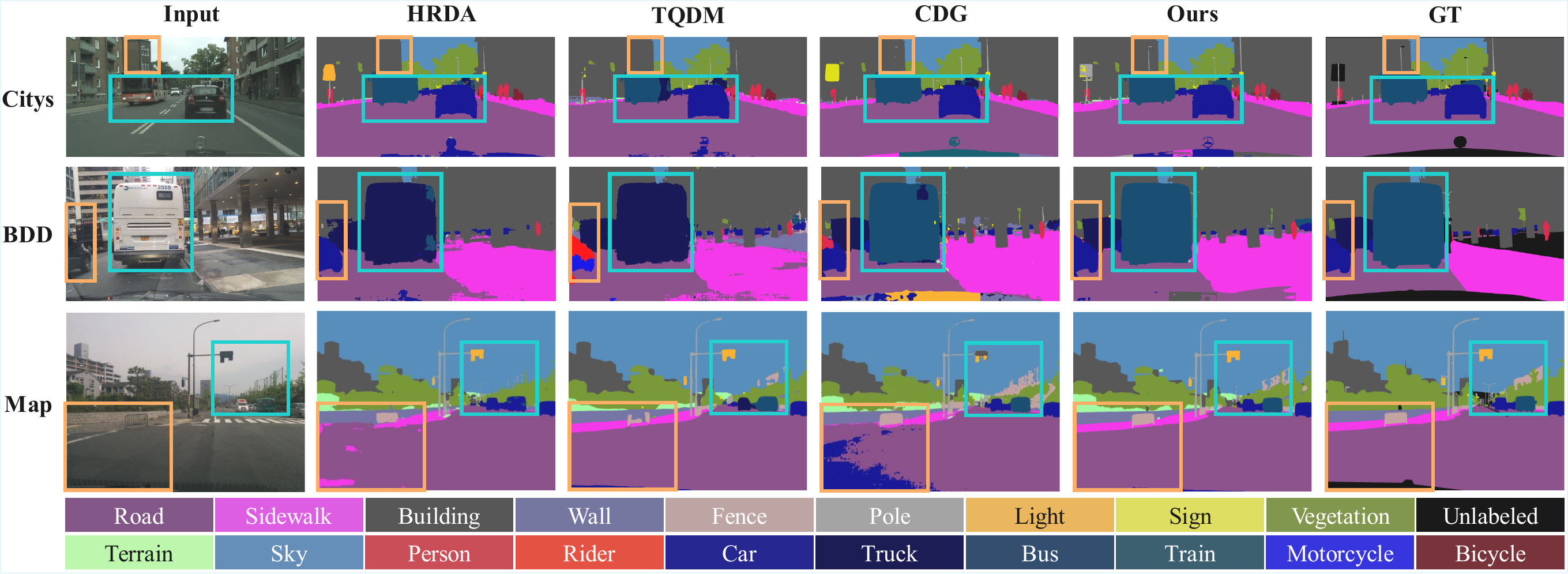}
    \caption{Qualitative comparison under the generalized multi-domain setting: \textit{GTAV $\rightarrow$ \{Cityscapes, BDD100K, Mapillary\}}. Rein-G achieves superior segmentation precision, notably: 1) Successfully detecting the lamppost in the highlighted area (row 1); 2) Producing the most accurate vehicle segmentation without misclassifying the motorcyclist as shown in row 2; 3) Simultaneously segmenting correct traffic signage and road surface in row 3.}
    \label{fig:qualitative}    
\end{figure*}

\begin{table*}[!htp]
\renewcommand\arraystretch{1.1}
\newcommand{\first}[1]{{\textbf{#1}}}
\newcommand{\second}[1]{{\underline{#1}}}
\caption{Ablation Study about {Rein} under \textit{GTAV $\rightarrow$ Cityscapes} generalization in terms of mIoU. Models are fine-tuned on GTAV and tested on Cityscapes. Components are progressively integrated to evaluate their benefits.} 
\label{tab:ablation_dg}%
\setlength\tabcolsep{3.0pt}
\resizebox{\linewidth}{!}{%
\begin{tabular}{l|l|c|ccccccccccccccccccc|>{\columncolor{gray!10}}c}
\hline
 Backbone  & \begin{tabular}[l]{@{}l@{}}Fine-tune\\ Method\end{tabular} & \begin{tabular}[c]{@{}c@{}}Trainable\\ Params$^*$\end{tabular} & Road&S.walk&Build.&Wall&Fence&Pole&Tr.Light&Sign&Vege.&Terrain&Sky&Person&Rider&Car&Truck&Bus&Train&M.bike&Bike&mIoU \\
 \hline
\multirow{6}{*}{\begin{tabular}[c]{@{}l@{}}DINOv2\\(Large)\\~\cite{Dinov2}\end{tabular}}& Full & 304.20M &92.5&57.4&90.5&52.7&50.5&55.0&67.2&48.9&90.4&43.4&\second{91.5}&79.5&50.3&92.3&50.0&71.1&61.3&46.9&56.9&65.0\\
 \cline{2-23}
  & Freeze  & ~~~0.00M&\second{93.6}&\second{62.8}&91.2&54.8&51.1&58.4&66.3&56.1&90.6&\first{47.7}&\first{91.8}&80.6&\first{53.1}&93.2&66.6&77.3&35.0&\first{64.7}&\second{59.5}&68.0\\
  & $+$ Rein-core  & ~52.84M &92.4&62.0&90.5&56.4&53.5&59.0&66.6&56.0&90.1&45.2&90.6&80.4&51.4&93.0&68.4&\first{88.5}&\first{71.1}&53.7&58.6&69.9\\
  & $+$ Rein-link & ~~~5.02M &93.0&61.8&\second{91.6}&\first{60.5}&\first{57.5}&58.5&\second{67.7}&58.7&\second{91.0}&44.6&90.7&\second{81.0}&51.3&\second{93.8}&66.5&\second{88.1}&67.1&52.3&56.3&70.1\\
  & $+$ Rein-lora & ~~~2.99M&93.0&62.3&\first{91.7}&57.3&55.5&\second{60.6}&\first{67.8}&58.4&{\first{91.1}}&{\second{46.3}}&91.4&80.6&49.4&93.6&69.2&86.8&\second{67.5}&51.8&57.1&70.1\\
  & $+$ Multi-head & ~~~2.99M&{\first{94.1}}&{\first{65.7}}&91.5&{\second{58.6}}&54.9&59.8&67.0&\second{58.8}&{\first{91.1}}&{46.3}&{\first{91.8}}&80.3&49.8&93.7&\second{69.8}&{86.8}&{67.1}&58.4&59.5&\second{70.8} \\
  & $+$ GELU & ~~~2.99M&93.1&61.9&{\first{91.7}}&57.0&{\second{56.4}}&{\first{61.5}}&{67.3}&{\first{60.3}}&{\first{91.1}}&44.0&89.9&{\first{81.1}}&{\second{52.5}}&{\first{94.0}}&{\first{72.0}}&85.8&65.9&{\second{58.9}}&{\first{62.7}}&{\first{70.9}} \\
  
 \hline
\end{tabular}
 }
\end{table*}

\section{Experiments}
\label{sec:experiments}

\subsection{Domain Generalization Semantic Segmentation (DGSS)}
\label{sec:DGSS}

To comprehensively validate the effectiveness of Rein-G, we conduct extensive experiments structured around three key aspects. First, we demonstrate that Rein-G exhibits superior generalization capabilities over state-of-the-art DGSS approaches (Table~\ref{tab:comparison_dg}). Second, a systematic evaluation across five leading Visual Foundation Models (VFMs) reveals that Rein-G outperforms alternative fine-tuning strategies (Table~\ref{tab:comparison_dg2}). Finally, to ensure a fair comparison, we establish Rein's superiority over both PEFT methods and existing DGSS approaches under an identical backbone architecture (Table~\ref{tab:comparison_dg2}, top). The subsequent section details the VFMs and experimental settings employed in these evaluations.

\subsubsection{The details of VFMs}
Our evaluation encompasses five distinct VFMs across three scales (Large, Giant, and 6B) to rigorously assess the broad applicability and scalability of Rein-G. This selection deliberately covers a diverse range of VFMs: language-image pre-training (CLIP\cite{CLIP}), hybrid language-image and masked image modeling (EVA02\cite{EVA,EVA02}), self-supervised learning on curated data (DINOv2\cite{Dinov2}), knowledge distillation from multiple VFMs (Radio\cite{radio}), and the vision encoder of a large vision-language model (InternVL-C~\cite{internvl_c}). To specifically evaluate scalability, we employ models of varying sizes: CLIP (Large), EVA02 (Large), DINOv2 (Large and Giant), Radio (Giant), and InternVL-C (6B). For comparison, we establish two fundamental baselines: (i) Full Fine-tuning, where all model parameters are updated, and (ii) Freeze, where the VFM backbone remains frozen and only the segmentation head is trained.

\subsubsection{Experiment settings}
Our experimental protocol is designed to assess generalization across two primary scenarios: \textbf{synthetic-to-real} and \textbf{real-to-real}.

For the synthetic-to-real scenario, we employ the following source datasets:
\begin{itemize}
    \item \textbf{GTAV}~\cite{gtav}: A large-scale synthetic dataset with 24,966 labeled images extracted from a video game.
    \item \textbf{Synthia}~\cite{SYNTHIA}: A collection of 9,400 photo-realistic frames ($1280 \times 960$) rendered from multiple viewpoints within a virtual city.
    \item \textbf{UrbanSyn}~\cite{gómez2023one}: A photo-realistic dataset of 7,539 images generated from semi-procedural synthetic urban driving scenarios.
\end{itemize}

The real-world target domains are represented by a diverse set of five datasets:
\begin{itemize}
    \item \textbf{Cityscapes (Citys)}~\cite{cityscapes}: A standard autonomous driving benchmark comprising 2,975 training and 500 validation images ($2048 \times 1024$).
    \item \textbf{BDD100K (BDD)}~\cite{bdd100k}: A diverse driving dataset from which we use its 1,000 validation images ($1280 \times 720$).
    \item \textbf{Mapillary Vistas (Map)}~\cite{mapillary}: A large-scale street-level imagery dataset, providing a validation set of 2,000 images ($1920 \times 1080$).
    \item \textbf{ACDC}~\cite{acdc}: A dataset specifically designed for evaluating robustness under adverse conditions, featuring images equally distributed across fog, nighttime, rain, and snow.
    \item \textbf{Cityscapes-C (Citys-C)}~\cite{hendrycks2018benchmarking,michaelis2019dragon}: A benchmark derived from Cityscapes by applying 16 diverse types of algorithmic corruption to test model robustness.
\end{itemize}

\subsubsection{Implementation details}
\label{sec:implementation}
We utilize the MMSegmentation ~\cite{mmseg2020} codebase for our implementation. For superior performance, mask2former~\cite{mask2former}, a widely-used segmentation head, is integrated with various VFMs that serve as the backbone. For the training phase, the AdamW optimizer~\cite{adamw} is employed, setting the learning rate at 1e-5 for the backbone and 1e-4 for both the decode head and the proposed Rein. Aiming to efficient training process, we utilize a configuration of 40,000 iterations with a batch size of 4, and crop images to a resolution of $512\times512$. Our approach includes only basic data augmentation, following Mask2Former~\cite{mask2former}. Thanks to our streamlined training configuration and reduced number of trainable parameters, \textbf{Rein-G can fine-tune models like DINOv2-Large or EVA02-Large on a single RTX 3090Ti GPU within 12 hours} for superior generalization ability. 
Note that, to achieve better performance and to align with the settings in \cite{tqdm} and \cite{TLDR}, we adjust the test resolution of Cityscapes to 2048×1024. As a result, the performance of the same methods reported in this paper may be slightly higher than that in our conference version.

\begin{table}[ht]
\centering
\renewcommand\arraystretch{1.1}
\setlength\tabcolsep{3.5pt}
\caption{Ablation study on lora dim $r$.}
\label{tab:ablation_dg_lora}
\begin{tabular}{ll|ccccc}
\hline
\multicolumn{2}{l|}{Rank $r$}& 4& 8& 16 & 32 & 64\\
\multicolumn{2}{l|}{Params}& 2.67M& 2.77M& 2.99M& 3.42M& 4.28M \\
\hline
\multirow{4}{*}{\begin{tabular}[c]{@{}l@{}}DINOv2\\(Large)~\cite{Dinov2}\end{tabular}} & Citys& 68.9 & 69.2 & 70.7 & 69.2 & 69.5 \\
 & BDD& 63.3 & 63.4 & 61.1& 63.8 & 64.1 \\
 & Map& 68.3 & 68.2 & 70.8 & 69.5 & 68.1 \\
\cline{2-7}
\rowcolor{gray!8}
 & Avg. & 66.8 & 67.0 & \textbf{67.5} & \textbf{67.5} & 67.2 \\

\hline
\end{tabular}

\end{table}
\begin{table}[!tbp]
\centering
\caption{Training Time, GPU Memory, and Storage of Domain Generalization Semantic Segmentation. All experiments were conducted on an NVIDIA A100 GPU with 40GB memory. The symbol \textbf{N/A} indicates that the corresponding model failed to train due to out-of-memory (OOM) errors.}
\label{tab:training_efficieny}
\renewcommand\arraystretch{1.1}
\setlength\tabcolsep{3.5pt}
\begin{tabular}{l|lc|ccc}
\hline
\multirow{2}{*}{VFMs} &
\multirow{2}{*}{Method} &
\multirow{2}{*}{\begin{tabular}[c]{@{}c@{}}Batch Size\end{tabular}} &
\multirow{2}{*}{\begin{tabular}[c]{@{}c@{}}Training\\Time\end{tabular}} &
\multirow{2}{*}{\begin{tabular}[c]{@{}c@{}}GPU\\Memory\end{tabular}} &
\multirow{2}{*}{\begin{tabular}[c]{@{}c@{}}Storage of \\ Trainable Parameters\end{tabular}}\\
&&&&\\
\hline
& Full &4& 11.2 h & 14.7 GB & 1.22 GB \\
\rowcolor{gray!8}
\multirow{-2}{*}{\begin{tabular}[c]{@{}c@{}} DINOv2\cellcolor{white}\\(Large)\cellcolor{white}\end{tabular}} & Rein-G &4& ~~\textbf{9.3} h & \textbf{9.9} GB & \textbf{0.09} GB \\
\hline
& \multirow{2}{*}{Full}&2& 16 h & 25.2 GB & \multirow{2}{*}{4.35GB} \\
&  &4& N/A & N/A & \\
\rowcolor{gray!8}
\multirow{-3}{*}{\begin{tabular}[c]{@{}c@{}} DINOv2\cellcolor{white}\\(Giant)\cellcolor{white}\end{tabular}} \cellcolor{white}& Rein-G &4& ~~\textbf{21.4} h & \textbf{24.7} GB & \textbf{0.11} GB \\
\hline
\end{tabular}%
\end{table}
\begin{figure*}
    \centering
    \includegraphics[width=0.9\linewidth]{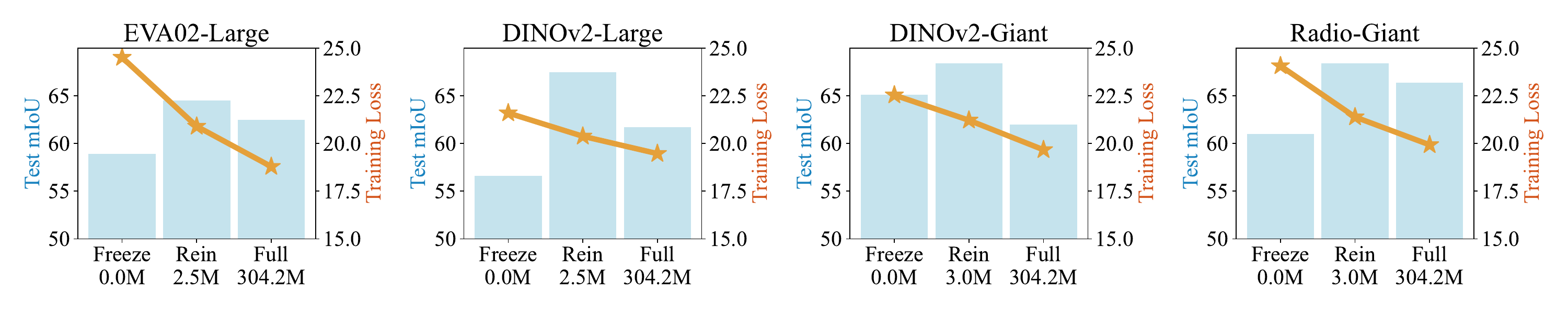}
    \caption{Analysis of the number of trainable parameters. The \textcolor{blue}{blue} bar charts in the figure represent the average mIoU tested on the Cityscapes, BDD100K, and Mapillary datasets, while the \textcolor{brown}{brown} line denotes the training loss during fine-tuning on GTAV dataset.}
    \label{fig:explain_parameters}
\end{figure*}
\begin{table}[htbp]
\centering
\caption{Ablation study on token length $m$.}
\label{tab:ablation_length}
\begin{tabular}{l|ccccc}
\toprule
Token length (m)         & 25   & 50   & 100  & 150  & 200  \\
\midrule
Trainable Parameters (M) &  2.96    & 2.97     &   2.99   &  3.01    & 3.03     \\
mIoU (\%)                & 70.0 & 70.2 & \textbf{70.7} & 70.6 & 68.9 \\
\bottomrule
\end{tabular}

\end{table}

\subsubsection{Comparison with SoTA DGSS Approaches}
\label{sec:sota_comparison}

As detailed in Table~\ref{tab:comparison_dg} and Table~\ref{tab:comparison_dg_c2cc_c2acdc}, Rein-G demonstrates consistent and significant performance gains across diverse domain generalization scenarios. 
In the synthetic-to-real setting, Rein-G paired with the Radio-Giant backbone achieves an average mIoU of 68.8\% on the \textit{GTAV $\rightarrow$ Citys.+BDD.+Map.} benchmark. This result represents a substantial improvement of 13.5 percentage points over ResNet101-based methods and surpasses the VFM-driven TQDM~\cite{tqdm} by 4.2\% mIoU on the Cityscapes validation set.
This advantage extends to multi-source settings, where Rein-G, utilizing the DINOv2-Giant backbone, establishes a new state-of-the-art under the \textit{GTAV+Synthia+UrbanSyn $\rightarrow$ Citys.+BDD.+Map.} configuration. 
Notably, in the challenging real-to-real generalization task (\textit{Citys. $\rightarrow$ ACDC(val)+BDD.+Map.}), the Radio-Giant variant of Rein-G attains 71.7\% mIoU, outperforming DGInStyle~\cite{dginstyle}, by a remarkable margin of 9.1\% mIoU. Visual results in Fig.\ref{fig:qualitative} further demonstrate enhanced boundary precision and class consistency.

\subsubsection{Comparison with the ``Full'' and ``Freeze'' Baselines}
\label{sec:baseline_comp}
As shown in Table~\ref{tab:comparison_dg2}, 
Rein-G demonstrates consistent advantages over both ``Full'' (complete parameter tuning) and ``Freeze'' (backbone frozen) baselines across five visual foundation models. The DINOv2-Giant implementation shows notable gains, surpassing ``Full'' baseline performance by 6.4\%, 3.9\%, and 2.4\% mIoU across three distinct evaluation settings. For extreme-scale models like InternVL-C 6B -- where full parameter tuning proves infeasible due to prohibitive memory demands (96GB GPU memory consumption even with batch size=1) -- Rein-G enables practical training through efficient parameter adaptation (24.0M trainable parameters). This parameter-efficient approach achieves 5.7\%, 4.1\%, and 4.6\% mIoU improvements over the ``Freeze'' baseline in respective scenarios, demonstrating both technical feasibility and performance superiority under hardware constraints.

\subsubsection{Comparison with the DGSS and PEFT methods under the same backbone}
The strong generalization capability of Rein-G arises not only from the powerful representation of VFMs, but also from its own effective architectural design. To quantitatively highlight the latter, we fix the backbone to Dinov2-Large and compare Rein-G with several representative PEFT and DGSS methods. Table~\ref{tab:comparison_dg2} reveals two key findings: (1) Frozen VFMs outperform full fine-tuning with specialized DGSS methods; (2) Rein-G achieves superior generalization, with a +2.3–5.8\% mIoU improvement over full fine-tuning, while introducing only 1\% additional parameters beyond the original backbone.

\begin{table*}
\centering
\setlength{\tabcolsep}{1.3pt}
\footnotesize
\belowrulesep=0pt
\aboverulesep=0pt
\renewcommand\arraystretch{1.1}
\definecolor{awardblue}{RGB}{0,0,147}   
\definecolor{awardgray}{RGB}{0,100,0} 
\definecolor{awardorange}{RGB}{204,0,0}  
\newcommand{\first}[1]{{\textbf{#1}}}
\newcommand{\second}[1]{{\underline{#1}}}
\newcommand{\third}[1]{\textcolor{awardgray}{\textbf{#1}}}
\caption{Performance Comparison with the proposed Rein against other \textbf{DASS} methods under three settings. Top-performing results are presented in \textbf{bold}, with \underline{second best} additionally emphasized.}
\label{tab:comparison_da}
\begin{tabular}{lc|ccccccccccccccccccc|c}
\toprule
Method&Publication&Road&S.walk&Build.&Wall&Fence&Pole&Tr.Light&Sign&Vege.&Terrain&Sky&Person&Rider&Car&Truck&Bus&Train&M.bike&Bike&mIoU\\
\midrule
\multicolumn{21}{c}{\textbf{Synthetic-to-Real:GTA$\to$Cityscapes(Val.)}}\\
\midrule
AdaptSeg~\cite{tsai2018learning}&CVPR18&85.2&20.4&85.5&38.2&30.9&34.5&43.0&26.2&87.4&40.3&86.4&63.6&23.7&88.6&48.5&50.6&5.8&33.1&16.2&47.8\\
DACS~\cite{tranheden2021dacs}&WACV21&88.9&50.0&88.4&46.4&43.9&43.1&53.4&54.8&89.9&{51.2}&92.8&64.2&9.4&91.4&77.3&63.3&0.0&47.4&49.8&58.2\\
ADPL~\cite{adpl}&TPAMI23&93.4&60.6&87.5&45.3&32.6&37.3&43.3&55.5&87.2&44.8&88.0&64.5&34.2&88.3&52.6&61.8&49.8&41.8&59.4&59.4\\
SePiCo~\cite{sepico}&TPAMI23&96.9&76.7&89.7&55.5&49.5&53.2&60.0&64.5&90.2&50.3&90.8&74.5&44.2&93.3&77.0&79.5&63.6&61.0&65.3&70.3\\
DAFormer~\cite{daformer-pami}&TPAMI23&95.7&70.2&89.4&53.5&48.1&49.6&55.8&59.4&89.9&47.9&92.5&72.2&44.7&92.3&74.5&78.2&65.1&55.9&61.8&68.3\\
HRDA~\cite{daformer-pami}&TPAMI23&{96.4}&{74.4}&{91.0}&{61.6}&{51.5}&{57.1}&{63.9}&{69.3}&{91.3}&48.4&{{{{94.2}}}}&{79.0}&{52.9}&{93.9}&{84.1}&{85.7}&{75.9}&{63.9}&{67.5}&{73.8}\\
MIC~\cite{mic}&CVPR23&{{{97.4}}}&{{{80.1}}}&91.7&61.2&56.9&59.7&66.0&{{{71.3}}}&{{\first{91.7}}}&51.4&{{{94.3}}}&79.8&56.1&94.6&85.4&90.3&80.4&64.5&{{68.5}}&75.9\\
MICDrop~\cite{micdrop}&ECCV24& {{\first{97.6}}}&{{\first{81.5}}}&92.0&62.8&59.4&{{\first{62.6}}}&62.9&{{\second{73.6}}}&{{\second{91.6}}}&52.6&94.1&80.2&57.0&94.8&87.4&90.7&81.6&65.3&67.8&76.6\\
HIAST~\cite{hiast}&TPAMI25&95.8&69.3&89.8&44.6&40.3&49.2&61.4&67.8&89.4&43.7&90.0&76.0&53.2&92.4&62.0&67.2&0.0&58.1&68.0&64.1\\
ICCLD~\cite{iccld}&CVPR25&\first{97.6}&74.3&90.9&62.3&52.3&57.0&64.9&72.5&91.1&51.3&\second{94.5}&78.6&53.2&\second{94.5}&84.9&85.4&75.1&65.4&66.7&74.4\\
QuadMix~\cite{quadmix}&TPAMI25&\second{97.5}&\second{80.9}&91.6&62.3&57.6&58.2&64.5&71.2&\first{91.7}&52.3&{94.3}&80.0&55.9&94.6&86.3&90.5&82.3&65.1&68.1&76.1\\
Yu \textit{et al.}~\cite{10891392}&TMM25&97.2&79.0&92.1&61.7&60.7&\second{62.2}&66.9&\first{74.2}&91.5&46.2&\first{95.1}&81.4&59.5&95.1&87.6&92.1&82.0&65.0&\first{70.2}&76.8\\
\midrule

\multicolumn{2}{l|}{Rein-A (CLIP-L)}&96.3&76.8&89.9&57.5&55.5&51.8&64.4&62.8&89.5&49.9&92.5&81.7&56.7&90.3&72.3&87.1&69.6&63.8&60.3&72.0\\

\multicolumn{2}{l|}{Rein-A (EVA02-L)}&96.8&75.0&91.6&63.7&{{\second{62.9}}}&56.5&68.1&65.4&90.7&49.3&90.9&82.7&64.0&\second{95.4}&{{\second{90.6}}}&93.4&87.4&67.2&66.5&76.7\\

\multicolumn{2}{l|}{Rein-A (DINOv2-L)}&93.2&63.4&{{\second{92.3}}}&{{\second{65.8}}}&61.5&60.3&71.0&66.1&91.5&53.5&93.8&\second{83.8}&65.9&\second{95.4}&89.6&93.7&85.8&68.1&65.7&76.9\\

\multicolumn{2}{l|}{Rein-A (DINOv2-G)}&91.2&57.8&{{92.2}}&{{\first{68.1}}}&{{\first{66.1}}}&{{61.4}}&{{\first{72.8}}}&66.4&91.2&52.7&93.3&{{\first{84.1}}}&\second{66.2}&{{\first{95.6}}}&88.9&{{\second{94.2}}}&88.0&67.2&65.9&{{\second{77.0}}}\\
\rowcolor{gray!10}
\multicolumn{2}{l|}{Rein-A (Radio-G)}&96.7&77.3&\first{92.5}&60.8&57.1&{61.9}&\second{71.5}&68.0&\first{91.7}&\first{55.6}&93.2&83.4&\second{66.2}&95.0&\first{91.4}&\first{94.8}&\first{89.4}&\second{71.3}&{68.7}&\first{78.2}\\
\multicolumn{2}{l|}{Rein-A (InternVL-6B)}&96.5&76.3&90.7&55.7&60.5&57.5&66.8&67.6&90.7&{{\second{53.7}}}&93.0&82.8&{{\first{66.8}}}&93.9&88.0&93.3&{{\second{88.5}}}&{{\first{72.3}}}&{{\second{69.3}}}&\second{77.0}\\
\midrule
\multicolumn{21}{c}{\textbf{Day-to-Adverse:Cityscapes$\to$ACDC(Test)}}\\
\midrule
ADVENT~\cite{vu2019advent}&CVPR19&72.9&14.3&40.5&16.6&21.2&9.3&17.4&21.2&63.8&23.8&18.3&32.6&19.5&69.5&36.2&34.5&46.2&26.9&36.1&32.7\\
GCMA~\cite{sakaridis2019guided}&ICCV19&79.7&48.7&71.5&21.6&29.9&42.5&56.7&57.7&{75.8}&39.5&{87.2}&57.4&29.7&80.6&44.9&46.2&62.0&37.2&46.5&53.4\\
MGCDA~\cite{sakaridis2020map}&TPAMI20&73.4&28.7&69.9&19.3&26.3&36.8&53.0&53.3&75.4&32.0&84.6&51.0&26.1&77.6&43.2&45.9&53.9&32.7&41.5&48.7\\
DANNet~\cite{wu2021dannet}&CVPR21&84.3&54.2&77.6&38.0&30.0&18.9&41.6&35.2&71.3&39.4&86.6&48.7&29.2&76.2&41.6&43.0&58.6&32.6&43.9&50.0\\
DAFormer~\cite{daformer-pami}&TPAMI23&58.4&51.3&84.0&42.7&35.1&50.7&30.0&57.0&74.8&52.8&51.3&58.3&32.6&82.7&58.3&54.9&82.4&44.1&50.7&55.4\\
HRDA~\cite{daformer-pami}&TPAMI23&{88.3}&{57.9}&{88.1}&{55.2}&{36.7}&{56.3}&{62.9}&{65.3}&74.2&{57.7}&85.9&{68.8}&{45.7}&{88.5}&{76.4}&{82.4}&{87.7}&{52.7}&{60.4}&{68.0}\\
MIC~\cite{mic}&CVPR23&90.8&67.1&89.2&54.5&40.5&57.2&62.0&68.4&76.3&61.8&87.0&71.3&49.4&89.7&75.7&86.8&89.1&56.9&63.0&70.4\\
CoDA~\cite{coda}&ECCV24&93.1&72.7&90.7&57.3&47.4&56.8&69.9&70.0&\first{87.3}&59.8&\first{95.4}&71.4&47.6&90.3&77.1&83.8&89.1&54.7&64.1&72.6\\
Yu \textit{et al.}~\cite{10891392}&TMM25&62.8&51.6&83.0&34.7&35.0&52.1&30.1&56.4&73.0&55.9&60.9&62.7&33.8&80.2&59.5&58.5&81.8&47.5&52.3&56.4\\
\midrule
\multicolumn{2}{l|}{Rein-A (CLIP-L)}&91.8&68.6&90.7&60.0&47.8&58.6&72.0&72.3&\second{86.9}&62.9&\second{94.6}&74.5&55.2&88.7&66.5&70.8&84.0&63.5&58.5&72.0\\

\multicolumn{2}{l|}{Rein-A~(EVA02-L)}&93.3&75.7&91.9&62.3&50.3&61.9&70.4&71.4&77.3&\first{69.8}&86.2&77.6&60.4&91.3&\second{80.7}&70.9&90.1&65.7&70.1&74.6\\

\multicolumn{2}{l|}{Rein-A~(DINOv2-L)}&\second{94.2}&{75.5}&{\second{92.0}}&{\first{69.3}}&\second{54.9}&{62.2}&{\first{79.0}}&{74.8}&{82.3}&\second{66.1}&{91.5}&{\first{81.7}}&{\second{64.3}}&{93.0}&{80.5}&{85.2}&{92.1}&{66.9}&{70.0}&{77.7}\\
\rowcolor{gray!10}
\multicolumn{2}{l|}{Rein-A~(DINOv2-G)}&\first{94.5}&\first{77.7}&91.8&\second{68.1}&\first{56.3}&\second{66.1}&\second{77.9}&\first{78.7}&82.4&65.8&91.6&79.7&58.3&\second{93.3}&\first{83.2}&\second{89.4}&\first{93.9}&\second{70.2}&\second{70.9}&\first{78.4}\\

\multicolumn{2}{l|}{Rein-A~(Radio-G)}&94.0&\second{76.5}&\first{92.4}&67.4&54.7&\first{66.8}&76.7&\second{77.7}&76.6&65.2&85.8&\second{81.0}&\first{65.4}&\first{93.8}&\first{83.2}&\first{91.5}&\second{93.8}&\first{72.7}&\first{71.8}&\second{78.3}\\

\multicolumn{2}{l|}{Rein-A (InternVL-6B)}&94.0&\second{76.5}&\first{92.4}&67.4&54.7&\first{66.8}&76.7&\second{77.7}&76.6&65.2&85.8&\second{81.0}&\first{65.4}&\first{93.8}&\first{83.2}&\first{91.5}&\second{93.8}&\first{72.7}&\first{71.8}&\second{78.3}\\
\midrule

\multicolumn{20}{c}{\textbf{Day-to-Nighttime:Cityscapes$\to$DarkZurich(Test)}}\\
\midrule
ADVENT~\cite{vu2019advent}&CVPR19&85.8&37.9&55.5&27.7&14.5&23.1&14.0&21.1&32.1&8.7&2.0&39.9&16.6&64.0&13.8&0.0&58.8&28.5&20.7&29.7\\
GCMA~\cite{sakaridis2019guided}&ICCV19&81.7&46.9&58.8&22.0&20.0&41.2&40.5&41.6&64.8&31.0&32.1&53.5&47.5&75.5&39.2&0.0&49.6&30.7&21.0&42.0\\
MGCDA~\cite{sakaridis2020map}&TPAMI20&80.3&49.3&66.2&7.8&11.0&41.4&38.9&39.0&64.1&18.0&55.8&52.1&53.5&74.7&66.0&0.0&37.5&29.1&22.7&42.5\\
DANNet~\cite{wu2021dannet}&CVPR21&90.0&54.0&{74.8}&{41.0}&{21.1}&25.0&26.8&30.2&\second{72.0}&26.2&{84.0}&47.0&33.9&68.2&19.0&0.3&66.4&38.3&23.6&44.3\\
DAFormer~\cite{daformer-pami}&TPAMI23&{93.5}&{65.5}&73.3&39.4&19.2&53.3&44.1&{44.0}&59.5&{34.5}&66.6&53.4&52.7&82.1&52.7&9.5&89.3&50.5&38.5&53.8\\
SePiCo~\cite{sepico}&TPAMI23&93.2&68.1&73.7&32.8&16.3&54.6&49.5&48.1&74.2&31.0&\second{86.3}&57.9&50.9&82.4&52.2&1.3&83.8&43.9&29.8&54.2\\
HRDA~\cite{daformer-pami}&TPAMI23&90.4&56.3&72.0&39.5&19.5&{57.8}&{52.7}&43.1&59.3&29.1&70.5&{60.0}&{58.6}&{84.0}&{\second{75.5}}&{11.2}&{90.5}&{51.6}&{40.9}&{55.9}\\
MIC~\cite{mic}&CVPR23&94.8&75.0&84.0&55.1&28.4&62.0&35.5&52.6&59.2&\second{46.8}&70.0&65.2&61.7&82.1&64.2&18.5&91.3&52.6&44.0&60.2\\
Yu \textit{et al.}~\cite{10891392}&TMM25&94.5&73.2&78.6&50.2&23.2&53.8&38.8&42.8&57.9&39.6&65.5&55.1&52.3&81.2&44.9&6.6&90.0&49.7&38.0&54.5\\
\midrule

\multicolumn{2}{l|}{Rein-A~(DINOv2-L)}&{\second{96.5}}&{\second{81.0}}&{\second{88.2}}&{\second{60.0}}&{\second{39.2}}&{\second{65.3}}&\second{59.8}&{\second{69.0}}&{\first{82.2}}&{41.7}&{\first{92.5}}&{\second{73.4}}&{\second{70.0}}&{\second{92.2}}&{\first{82.0}}&\second{45.9}&{\second{94.5}}&{\second{70.0}}&\second{44.7}&{\second{70.9}}\\

\rowcolor{gray!10}
\multicolumn{2}{l|}{Rein-A~(DINOv2-G)}&\first{99.4}&\first{85.1}&\first{92.3}&\first{62.6}&\first{48.9}&\first{69.3}&\first{68.1}&\first{79.7}&52.9&\first{47.0}&51.2&\first{77.8}&\first{74.6}&\first{95.1}&73.2&\first{54.4}&\first{98.4}&\first{76.2}&\first{51.4}&\first{71.5}\\
\bottomrule
\end{tabular}

\end{table*}

\subsubsection{Analysis of Fewer Trainable Parameters}
Classical neural network theory~\cite{delemma,hastie2009elements} points out that as model capacity increases, the empirical risk (or training risk) monotonically decreases, indicating an improved fit to training data. Conversely, the true risk (or test risk) typically exhibits a ``U-shaped" curve, initially decreasing and then increasing, a phenomenon known as overfitting. From a modern viewpoint, the scaling law~\cite{scaling} suggests that on a smaller fixed dataset, performance stops to improve as model parameters increase, leading to overfitting.

In the majority of general tasks, the practice of early-stopping, based on evaluation data, can partly mitigate overfitting. However, in the field of domain generalization, the unknown test data distribution makes acquiring a valid evaluation dataset unavailable. Moreover, fine-tuning datasets are often smaller compared to ImageNet~\cite{imagenet} or LVD-142M~\cite{Dinov2}. Hence, employing fewer trainable parameters emerges as a strategic approach to mitigate overfitting.

In Sec.~\ref{sec:DGSS}, extensive experiments comprehensively demonstrate Rein-G's pivotal role in enhancing the generalization capabilities of VFMs. This enhancement may be attributed to two factors: 1) Rein's improved fitting capability for VFMs, ensuring better alignment with training data; 2) Rein-G's reduction of overfitting in VFMs during fine-tuning on smaller datasets, thus exhibiting enhanced generalization in testing. To delve into this, we analyze and compare the average training loss in the final 1000 iterations of the fine-tuning phase and their corresponding test metrics for various VFMs and decode heads.

Fig.~\ref{fig:explain_parameters} showcases a consistent trend across four different configurations. Intuitively, as trainable parameters increase from $0.00M (Freeze)\rightarrow 2.53M (Rein-G) \rightarrow 304.24M (Full)$, the training loss monotonically decreases, indicating that a greater number of trainable parameters indeed better fit the training dataset. However, the test metrics on the target dataset peak with Rein-G, which employs 2.53 million parameters and incurs a sub-optimal training loss. In contrast, the ``Full" baseline, despite recording the lowest training loss, only achieves sub-optimal test performance, a clear indicator of overfitting when compared to other setups. This observation aligns with the conclusions in ~\cite{delemma,scaling}, supporting our observation that fewer trainable parameters can lead to superior generalizability.

\subsubsection{Component Ablation Studies} 
Table~\ref{tab:ablation_dg} is dedicated to thoroughly examining the effectiveness of each component within Rein. In the \textit{GTAV $\rightarrow$ Citys} generalization setting, we sequentially incorporate different components of Rein-G and assess their impact. Interestingly, we observe that the ``Freeze" occasionally exhibits better recognition for specific categories, e.g., `road, sidewalk', compared to the ``Full". This suggests that VFMs lose some pre-training knowledge during fine-tuning, and ``Freeze" helps to prevent. Similarly, our methods mitigate this knowledge forgetting. Furthermore, our methods show improved recognition capabilities for the majority of the 19 categories. For example, in recognizing `wall, motorcycle, bicycle', our approach significantly outperforms both the ``Full" and ``Freeze" baselines.

Overall, ``Rein-core" boosts the average performance across 19 classes. Furthermore, ``Rein-link" further boosts accuracy for certain objects, including `car, bus, train, motorcycle', especially for DINOv2. Employing layer-shared MLP weights and low-rank token sequence efficiently reduce the number of trainable parameters and positively influences the performance of the model.

\subsubsection{Speed, memory, and storage} 
For practical applications, training speed, GPU memory usage, and model storage requirements are crucial. As shown in Table~\ref{tab:training_efficieny}, compared to ``Full" baseline, proposed Rein-G improves training speed and reduces GPU memory usage. A significant advantage of Rein-G is that models trained under different settings can share the same backbone parameters. This means that for switch in diverse tasks and settings, we can only store and swap the Rein-G weights (0.01GB) and head weights (0.08GB), rather than all parameters. Moreover, during training DINOv2-Giant, Rein-G enables the use of larger batch sizes which are impractical under full fine-tuning.

\subsubsection{Analysis on token length $m$}
The core component of Rein-G is learnable tokens $T\in\mathbb{R}^{m\times c}$. We explored various lengths $m$ for the token sequence, ranging from 25 to 200. As demonstrated in Tab.~\ref{tab:ablation_length}, models with $m=100$ achieve a strong mIoU of 70.7\%. We ultimately selected $m=100$ as the most suitable parameter.

\subsubsection{Analysis on rank $r$}
As shown in Table~\ref{tab:ablation_dg_lora}, we turn attention to the effect of rank $r$ on model performance. With DINOv2 as the backbone, the optimal results are observed at $r=16$ and $r=32$. Consequently, unlike LoRA~\cite{lora}, we opt for a comparatively higher value of $r=16$ for our model.

\begin{table*}
\centering
\setlength{\tabcolsep}{1.3pt}
\footnotesize
\belowrulesep=0pt
\aboverulesep=0pt
\renewcommand\arraystretch{1.1}
\definecolor{awardblue}{RGB}{0,0,147}   
\definecolor{awardgray}{RGB}{0,100,0} 
\definecolor{awardorange}{RGB}{204,0,0}  
\newcommand{\first}[1]{{\textbf{#1}}}
\newcommand{\second}[1]{{\underline{#1}}}
\newcommand{\third}[1]{\textcolor{awardgray}{{#1}}}
\caption{Ablation Study about {MCAS} in terms of mIoU. Components are removed to evaluate their benefits.}
\label{tab:ablation_da}
\begin{tabular}{l|ccccccccccccccccccc|c}
\toprule
Method&Road&S.walk&Build.&Wall&Fence&Pole&Tr.Light&Sign&Vege.&Terrain&Sky&Person&Rider&Car&Truck&Bus&Train&M.bike&Bike&mIoU\\
\midrule
\rowcolor{gray!10}
Rein-A&{\second{93.2}}&63.4&{{\first{92.3}}}&{{\first{65.8}}}&{{\first{61.5}}}&{\second{60.3}}&{{\first{71.0}}}&{\second{66.1}}&{{\first{91.5}}}&{{\first{53.5}}}&{{\first{93.8}}}&{{\first{83.8}}}&{{\first{65.9}}}&{{\first{95.4}}}&{{\first{89.6}}}&{{\first{93.7}}}&{{\second{85.8}}}&68.1&{{\first{65.7}}}&{{\first{76.9}}}\\
w/o Rein-G&93.2&{{\first{73.4}}}&89.8&59.4&{\second{60.3}}&57.6&45.9&{{\first{74.3}}}&89.2&49.9&90.1&81.7&63.6&93.3&87.3&91.6&84.5&{{\first{72.4}}}&0.3&71.5\\
w/o $\mathcal{L}_{mix}$&{{\first{95.5}}}&{\second{68.8}}&91.3&55.1&57.1&58.0&67.2&60.4&90.7&41.5&91.6&79.5&53.4&91.8&54.9&90.6&{85.4}&63.2&{\second{65.3}}&71.6\\
w/o $\mathcal{L}_{mask}$&90.8&57.4&{\second{91.7}}&60.5&53.5&{{\first{60.5}}}&{\second{68.8}}&59.0&{\second{91.0}}&{\second{52.4}}&{92.2}&{\second{83.5}}&{64.2}&{\second{95.2}}&{\second{89.0}}&{91.7}&78.4&{\second{69.8}}&64.1&{\second{74.4}}\\
w/o $\mathcal{M}_{stm}$&89.5&55.9&90.9&{\second{63.3}}&59.6&59.0&68.6&65.7&90.1&52.3&\second{93.0}&82.5&\second{64.4}&93.7&80.5&\second{92.6}&\first{86.2}&53.8&60.9&71.8\\
\bottomrule
\end{tabular}

\end{table*}
\begin{table*}
\centering
\setlength{\tabcolsep}{1.3pt}
\footnotesize
\belowrulesep=0pt
\aboverulesep=0pt
\renewcommand\arraystretch{1.1}
\definecolor{awardblue}{RGB}{0,0,147}   
\definecolor{awardgray}{RGB}{0,100,0} 
\definecolor{awardorange}{RGB}{204,0,0}  
\newcommand{\first}[1]{{\textbf{#1}}}
\newcommand{\second}[1]{{\underline{#1}}}
\newcommand{\third}[1]{{\textbf{#1}}}
\caption{Ablation study on SAM. Models trained on GTAV and adapted to Cityscapes, evaluated on the val set. Best results \textbf{boldfaced}.}
\label{tab:ablation_sam}
\begin{tabular}{l|ccccccccccccccccccc|c}
\toprule
Model&Road&S.walk&Build.&Wall&Fence&Pole&Tr.Light&Sign&Vege.&Terrain&Sky&Person&Rider&Car&Truck&Bus&Train&M.bike&Bike&mIoU\\
\midrule
with 
SAM1~\cite{SAM}&\second{90.7}&\second{56.9}&\second{91.7}&\first{66.3}&\second{58.9}&\first{60.7}&\second{67.3}&\first{68.2}&\second{91.1}&\second{52.6}&\second{93.2}&\second{82.1}&\second{65.3}&\second{94.3}&\second{81.3}&\second{93.2}&\second{83.8}&\second{67.7}&\first{68.7}&\second{75.5}\\
\rowcolor{gray!10}
with 
SAM2~\cite{SAM2}&\first{93.2}&\first{63.4}&\first{92.3}&\second{65.8}&\first{61.5}&\second{60.3}&\first{71.0}&\second{66.1}&\first{91.5}&\first{53.5}&\first{93.8}&\first{83.8}&\first{65.9}&\first{95.4}&\first{89.6}&\first{93.7}&\first{85.8}&\first{68.1}&\second{65.7}&\first{76.9}\\
\bottomrule
\end{tabular}

\end{table*}
\begin{table}[!tbp]
\centering
\caption{Training Time, GPU Memory, and Storage Consumption for Domain Adaptation Semantic Segmentation.
All experiments were conducted on an NVIDIA A100 GPU with 40GB memory. The symbol \textbf{N/A} indicates that the corresponding model failed to train due to out-of-memory (OOM) errors.}
\label{tab:training_efficieny_da}
\renewcommand\arraystretch{1.1}
\setlength\tabcolsep{3.5pt}
\begin{tabular}{l|lc|ccc}
\hline
\multirow{2}{*}{VFMs} &
\multirow{2}{*}{Method} &
\multirow{2}{*}{Batch Size} &
\multirow{2}{*}{\begin{tabular}[c]{@{}c@{}}Training\\Time\end{tabular}} &
\multirow{2}{*}{\begin{tabular}[c]{@{}c@{}}GPU\\Memory\end{tabular}} &
\multirow{2}{*}{\begin{tabular}[c]{@{}c@{}}Storage of\\Trainable Parameters\end{tabular}}\\
&&&&\\
\hline
& Full && 25.2 h & 22.9 GB & 1.22 GB \\
\rowcolor{gray!10}
\multirow{-2}{*}{\begin{tabular}[c]{@{}c@{}} DINOv2\cellcolor{white}\\(Large)\cellcolor{white}\end{tabular}}& Rein-A &\multirow{-2}{*}{1}& \textbf{16.7} h & \textbf{15.0} GB & \textbf{0.09} GB \\
\hline
& Full && N/A & N/A & 4.35 GB \\
\rowcolor{gray!10}
\multirow{-2}{*}{\begin{tabular}[c]{@{}c@{}} DINOv2\cellcolor{white}\\(Giant)\cellcolor{white}\end{tabular}} &  Rein-A &\multirow{-2}{*}{1}& \textbf{30 h} & \textbf{34.9} GB & \textbf{0.11} GB \\
\hline
\end{tabular}%
\end{table}
\begin{table}[htbp]
\centering
\caption{Sensitivity analysis of loss weights $\alpha$ and $\beta$ on Cityscapes $\to$ ACDC adaptation task (validation mIoU (\%)}
\label{tab:alpha_beta_sensitivity}
\resizebox{\linewidth}{!}{
\begin{tabular}{cccccc}
\toprule
\multirow{2}{*}{$\alpha$} &  \multicolumn{5}{c}{Performance (mIoU)} \\
\cmidrule(lr){2-6}
 &  {$\beta=10^{-1}$} & {$\beta=10^{-0.5}$} & {$\beta=10^{0}$} & {$\beta=10^{0.5}$} & {$\beta=10^{1}$} \\
\midrule
0.1  & 74.3 & 74.5 & 74.3 & 70.7 & 70.5 \\
1  & 75.8 & 76.1 & \textbf{76.9} & 74.8 & 73.8 \\
10& 75.2 & 75.3 & 74.9 & 74.9 & 74.8 \\
\bottomrule
\end{tabular}
}
\end{table}

\subsection{Domain Adaptation Semantic Segmentation (DASS)}
In this section, we comprehensively evaluate Rein-A over four datasets within three adaptation settings: (1) \textit{GTAV $\rightarrow$ Cityscapes}, (2) \textit{Cityscapes $\rightarrow$ DarkZurich}, and (3) \textit{Cityscapes $\rightarrow$ ACDC}. Rein-A is benchmarked against state-of-the-art methods~\cite{daformer-pami,ddb,quadmix,iccld,hiast,micdrop,mic,sepico}. 

\subsubsection{Implementation Details}

Our implementation is built upon the MMSegmentation~\cite{mmseg2020} codebase. We integrate and re-implement key components from several state-of-the-art training pipelines, including DDB~\cite{ddb}, DACS~\cite{tranheden2021dacs}, MIC~\cite{mic}, and DAFormer~\cite{daformer-pami}. Specifically, our methodology incorporates:
\begin{itemize}
    \item Rare Class Sampling (RCS) from DAFormer~\cite{daformer-pami}.
    \item Data augmentations such as class-mix from DDB~\cite{ddb}.
    \item Mask-based image training inspired by MIC~\cite{mic}.
    \item An Exponential Moving Average (EMA) teacher model update from DACS~\cite{tranheden2021dacs}.
\end{itemize}
For pseudo-label generation and weighting, we adopt the strategy from DAFormer. During adaptation, the VFM backbone parameters remain frozen, while only the segmentation decoder and our proposed Rein-G module are fine-tuned. The Rein-G module is initialized by pre-training on the source domain data. Unless otherwise specified, all other hyperparameters and implementation details are consistent with those outlined in Section~\ref{sec:implementation}.

\subsubsection{Synthetic-to-Real Adaptation of Rein-A}
As detailed in Table~\ref{tab:comparison_da}, Rein-A demonstrates superior performance on the synthetic-to-real adaptation benchmark GTAV $\rightarrow$ Cityscapes. It consistently outperforms prominent methods such as DAFormer and HRDA~\cite{daformer-pami}. Notably, Rein-A achieves a significant 2.3 percentage point increase in mIoU over MIC~\cite{mic}, a strong contemporary baseline. We attribute this performance advantage to the powerful semantic representations of the VFM, which enables Rein-A to more accurately resolve ambiguities between challenging categories. For instance, as observed in the table, our method shows {comparable performance} on semantically simple categories like \texttt{road}, \texttt{sidewalk}, and \texttt{sky}, where most models already perform well. In contrast, for semantically complex classes such as \texttt{Person}, \texttt{Rider}, and the various vehicle types (\texttt{Car}, \texttt{Truck}, \texttt{Bus}, \texttt{Train}), which contain rich and fine-grained details, our model demonstrates a distinct advantage.

\subsubsection{Real-to-Real Adaptation of Rein-A}
The efficacy of Rein-A is further validated on challenging real-to-real adaptation tasks, namely Cityscapes $\rightarrow$ Dark Zurich and Cityscapes $\rightarrow$ ACDC, with results detailed in Table~\ref{tab:comparison_da}. Across both benchmarks, Rein-A consistently outperforms state-of-the-art methods, including DAFormer~\cite{daformer-pami} and MIC~\cite{mic}. This superiority is particularly pronounced when compared to MIC, where Rein-A achieves substantial mIoU gains of {8.0\%} on the Cityscapes $\rightarrow$ ACDC task and a remarkable {11.3\%} on the Cityscapes $\rightarrow$ Dark Zurich task.

\subsubsection{Ablation Study of Rein-A Components}
To dissect the contribution of each key component within Rein-A, we conduct a comprehensive ablation study on the GTAV $\rightarrow$ Cityscapes benchmark using the DINOv2-L backbone. The results, presented in Table~\ref{tab:ablation_da}, affirm that all components are integral to the model, as the removal of any single module leads to a significant performance decline. Specifically, ablating the Rein-G module (\textit{w/o Rein-G}) and reverting to a full fine-tuning approach results in a substantial 5.4\% drop in mIoU. Delving into the module's internal architecture, we find that while omitting the mask branch yields a 2.5\% mIoU reduction, ablating the mix branch incurs a more severe performance degradation, indicating its more critical role in the adaptation process. Finally, the removal of the STM module, which integrates knowledge from SAM, leads to a 5.1\% mIoU drop.

\subsubsection{Speed, memory, and storage}
Rein-A introduces substantial improvements in computational efficiency over standard full fine-tuning, significantly reducing both training time and GPU memory consumption, as detailed in Table~\ref{tab:training_efficieny_da}. This efficiency stems from its compact parameter space; Rein-A requires only 0.11\,GB for its trainable parameters, in stark contrast to the 4.35\,GB required for the full DINOv2-Giant backbone. This lightweight design facilitates the storage of multiple domain-specific checkpoints with minimal overhead, an advantage for multi-target adaptation scenarios.
Crucially, this memory efficiency addresses a major bottleneck in adapting large-scale models. For instance, fully fine-tuning the DINOv2-Giant model is infeasible on a standard 40\,GB NVIDIA A100 GPU, failing even with a batch size of one. By drastically reducing the memory footprint, Rein-A makes the adaptation of billion-parameter models not only possible but practical on widely available hardware.

\subsubsection{Sensitivity Analysis of Hyperparameters $\alpha$ and $\beta$}
We conduct a sensitivity analysis to evaluate the impact of the loss weights $\alpha$ and $\beta$ from Eq.~(\ref{eq:totalloss}) on performance. The results, presented in Table~\ref{tab:alpha_beta_sensitivity}, reveal that Rein-A is robust to variations in these hyperparameters. Most configurations yield strong performance, with mIoU falling within a narrow 74\% to 77\% range. While performance remains stable across a wide spectrum of values, some settings (e.g., $\alpha=0.1, \beta=10$) can lead to suboptimal outcomes. Peak performance is achieved when $\alpha = \beta = 1$; therefore, we adopt this configuration for all experiments reported in this paper.

\subsubsection{Analysis of SAM Model Choice}
We investigate the impact of employing different SAM models in Sec.~\ref{sec:stm}, with results presented in Table~\ref{tab:ablation_sam}. It is crucial to note that our framework leverages SAM exclusively during the training. Consequently, the choice of SAM model has no bearing on inference speed or the model's parameter count. Furthermore, the SAM-generated masks are pre-computed offline in a pre-processing step, thereby adding no computational overhead to the training loop.
Turning to the performance comparison, the results indicate that utilizing SAM2~\cite{SAM2} yields the highest adaptation performance. Based on this finding, we adopt SAM2 as the default configuration in our study.

\section{Conclusion}
In this paper, we have presented Rein++, a unified and highly effective framework for adapting Vision Foundation Models (VFMs) to both Domain Generalization and Unsupervised Domain Adaptation for semantic segmentation. We demonstrated that its first component, \textbf{Rein-G}, provides a parameter-efficient and scalable fine-tuning solution for DG, successfully handling models with up to six billion parameters. Building on this foundation, we introduced \textbf{Rein-A}, a novel self-training method for UDA that leverages a sophisticated dual-level alignment strategy to ensure stable and accurate adaptation without target labels.

The core strength of Rein++ lies in the synergy between these components: Rein-G's efficiency makes large-scale VFM adaptation feasible, while Rein-A's robust mechanism makes it effective. Our extensive experiments validate this, showing that Rein++ consistently establishes a new state of the art across multiple challenging DG and DA benchmarks. By making the adaptation of massive VFMs both practical and powerful, this work paves the way for their broader application in complex, real-world segmentation scenarios.
\bibliographystyle{IEEEtran}
\bibliography{main}

\begin{IEEEbiography}[{\includegraphics[width=1in,height=1.25in,keepaspectratio]{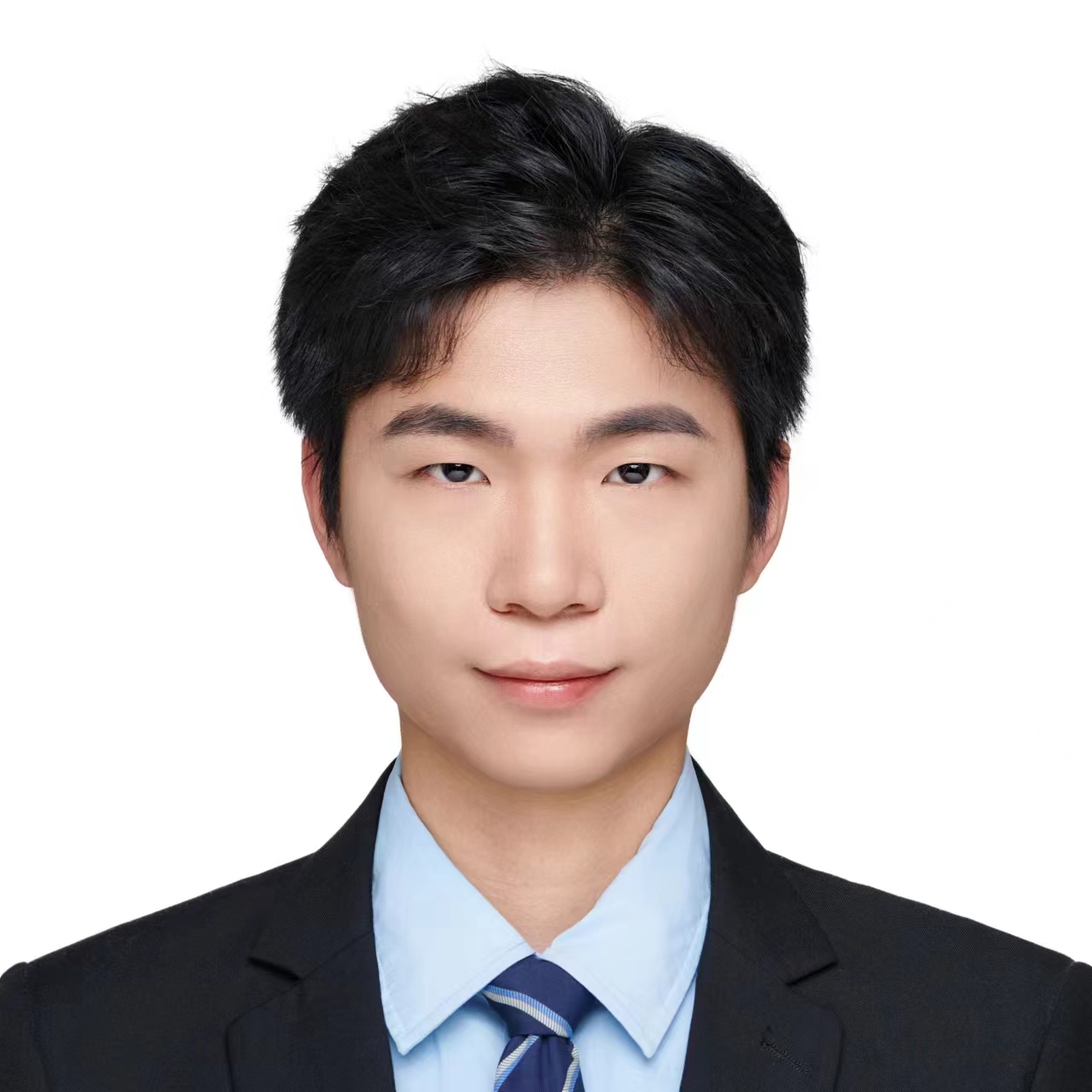}}]
    {Zixiang Wei} received the B.S. degree in Communication Engineering from the School of Electronic and Information Engineering, Anhui University, Hefei, China, in 2021. He is currently pursuing the Ph.D. degree in Instrument Science and Technology with the School of Engineering Science, University of Science and Technology of China, Hefei, China. His research interests include Domain Generalized Semantic Segmentation, Domain Adaptive Semantic Segmentation, and Vision Foundation Models.
\end{IEEEbiography}

\begin{IEEEbiography}[{\includegraphics[width=1in,height=1.25in,keepaspectratio]{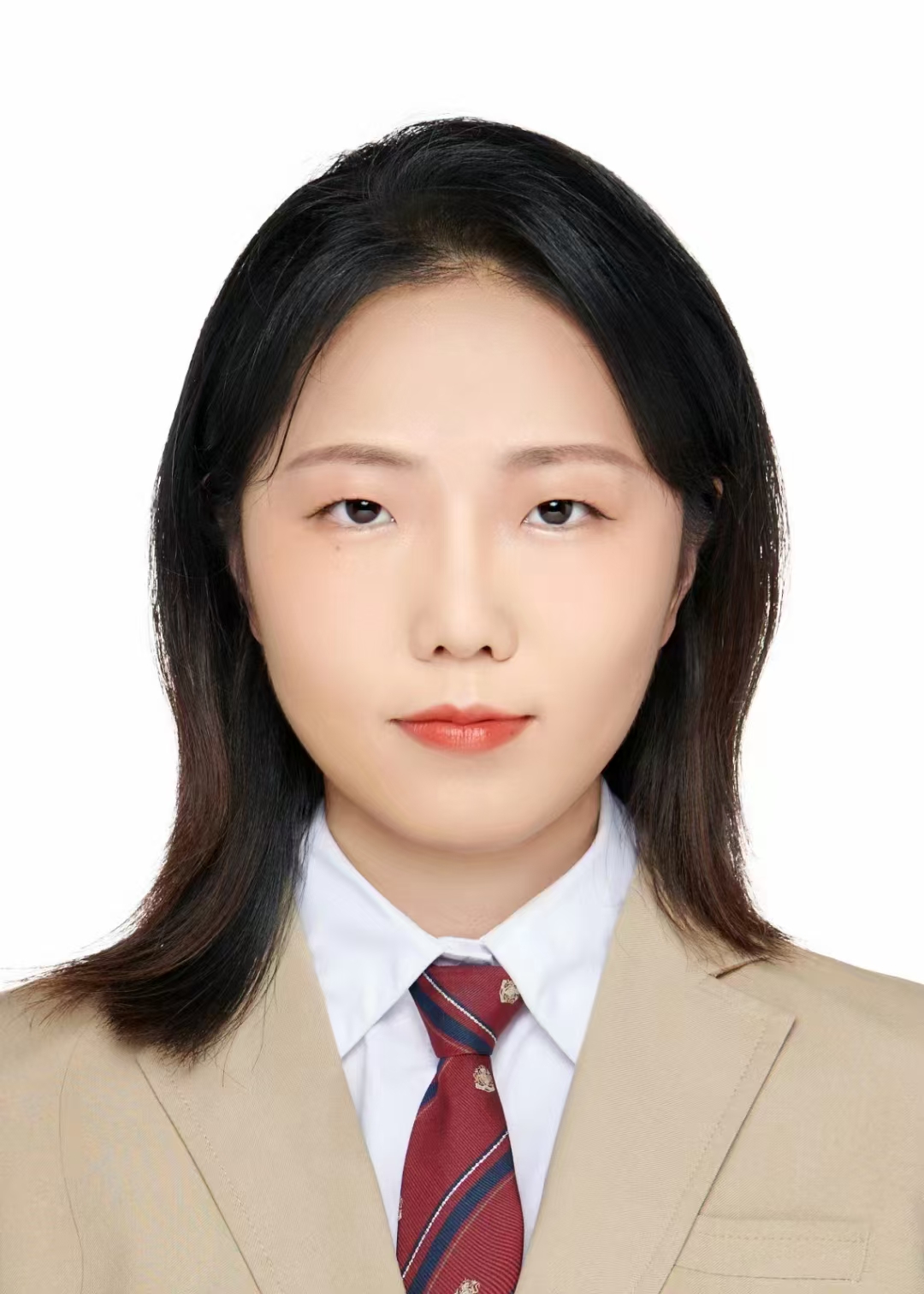}}]
    {Xiaoxiao Ma} received her B.S. degree in Computer Science and Technology from China Agricultural University, Beijing, China, in 2022. She is currently pursuing a Ph.D. degree in Automation at the School of Engineering Science, University of Science and Technology of China. Her research interests include multimodal learning and generative models.
\end{IEEEbiography}


\begin{IEEEbiography}[{\includegraphics[width=1in,height=1.25in,keepaspectratio]{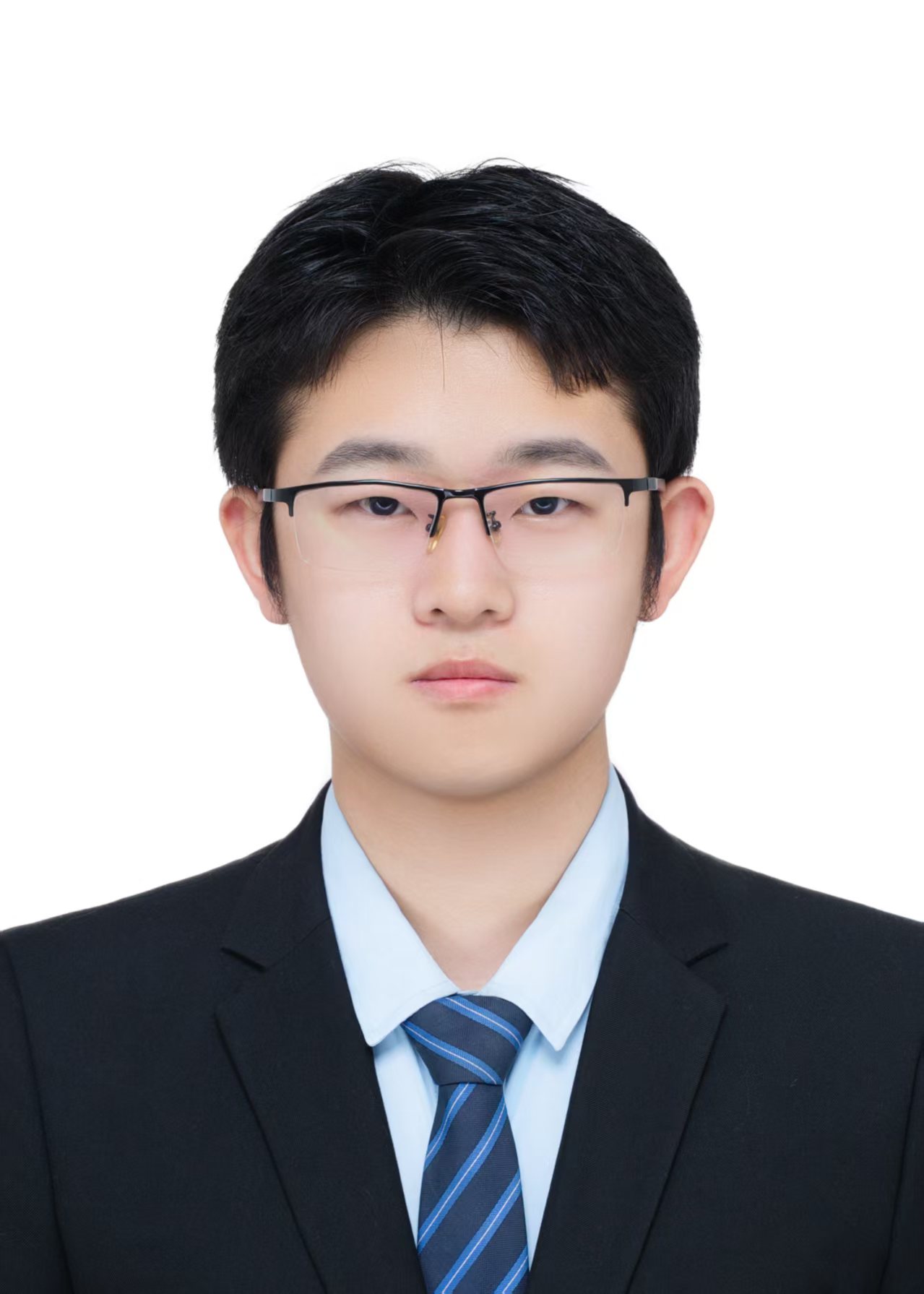}}]
    {Ruishen} Yan received the B.E. degree from Anhui University, Hefei, China, in 2025. He is currently pursuing the M.S. degree with the School of Engineering Science, University of Science and Technology of China. His research interests include deep learning, defect detection, and medical image segmentation.
\end{IEEEbiography}

\begin{IEEEbiography}[{\includegraphics[width=1in,height=1.25in,keepaspectratio]{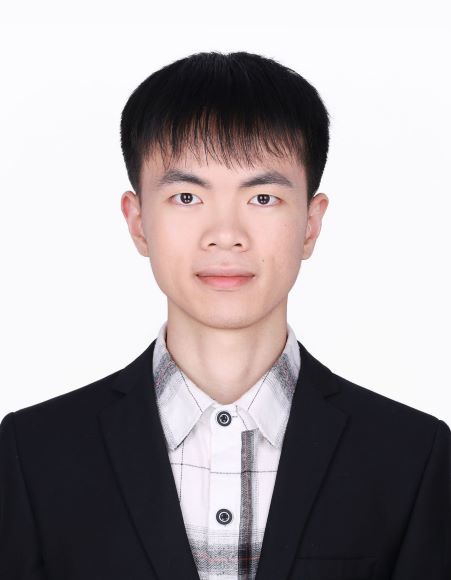}}]
{Tao Tu} received his B.S. degree in Communication Engineering from Anhui University, Hefei, China, in 2022. He is currently pursuing a Ph.D. in Instrument Science and Technology at the School of Engineering Science, University of Science and Technology of China, Hefei, China. His research interests focus on machine vision and FPGA acceleration.
\end{IEEEbiography}

\begin{IEEEbiography}[{\includegraphics[width=1in,height=1.25in,clip,keepaspectratio]{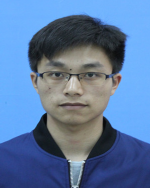}}]{Huaian Chen}
	received the Ph.D. degree from the University of Science and Technology of China, Hefei, China, in 2022. He is currently a Postdoctoral Research Fellow with the School of Engineering Science, University of Science and Technology of China. 	
	His research interests include deep learning, image/video segmentation, and image/video restoration.
\end{IEEEbiography}

\begin{IEEEbiography}[{\includegraphics[width=1in,height=1.25in,clip,keepaspectratio]{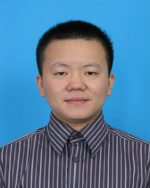}}]{Yi Jin}
	(Member, IEEE) received the Ph.D. degree from the University of Science and Technology of China, Hefei, China, in 2013. 
	He is currently an Associate Professor with the School of Engineering Science, University of Science and Technology of China. He has authored or coauthored over 50 refereed articles. His current research interests include intellectual detection, image processing, and artificial intelligence. 	
	Dr. Jin was a recipient of the Key Innovations Award of the Chinese Academy of Sciences and the First Class Science and Technology Progress Award of Anhui Province in 2016 and 2019.
\end{IEEEbiography}

\begin{IEEEbiography}[{\includegraphics[width=1in,height=1.25in,clip,keepaspectratio]{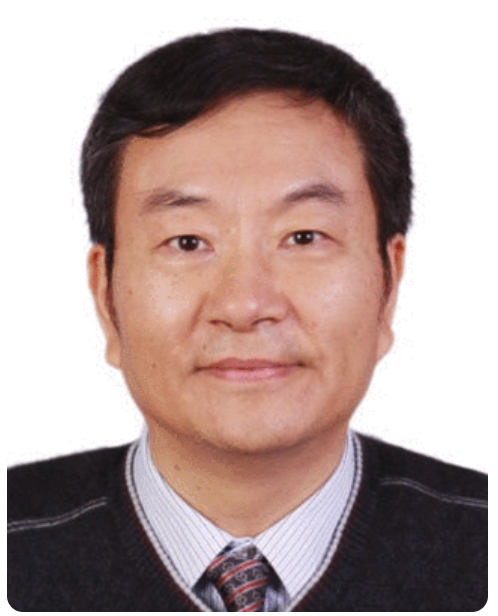}}]{Jinjin Zheng} received the Ph.D. degree in computer-aided geometric modeling from the University of Birmingham, Birmingham, U.K., in 1998.,He is currently a Professor with the School of Engineering Science, University of Science and Technology of China, Hefei, China. His research interests include computer-aided geometric design, and micro-electro-mechanical and computer simulation.
\end{IEEEbiography}

\begin{IEEEbiography}[{\includegraphics[width=1in,height=1.25in,clip,keepaspectratio]{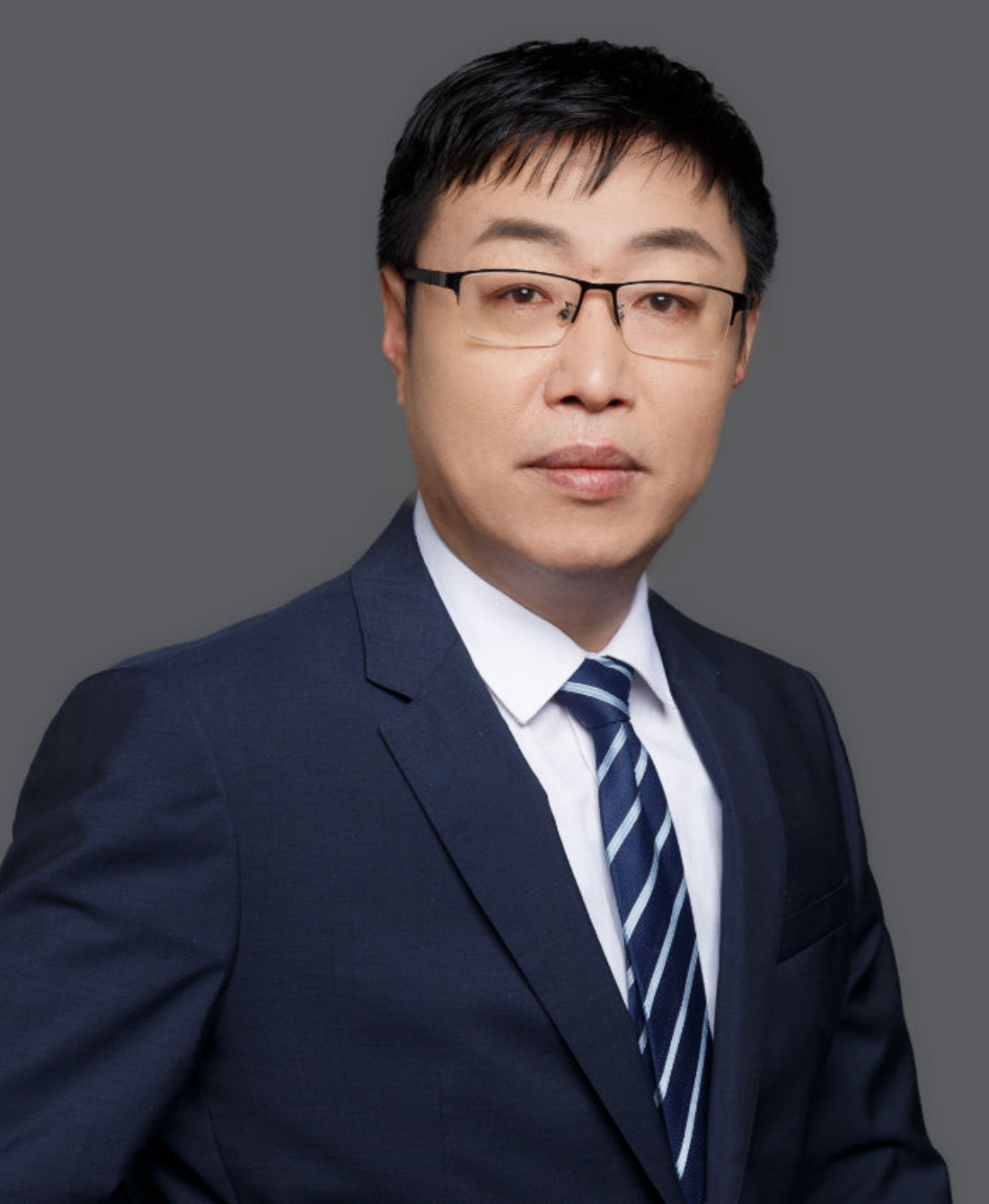}}]{Enhong Chen} (Fellow, IEEE) received the Ph.D. degree from the University of Science and Technology of China, Hefei, China. He is a professor and vice dean of the School of Computer Science, USTC. His general area of research include data mining and machine learning, social network analysis, and recommender systems. He has published more than 100 papers in refereed conferences and journals, including IEEE Transactions on Knowledge and Data Engineering, IEEE Transactions on Industrial Electronics, KDD, ICDM, NIPS, and CIKM. He was on program committees of numerous conferences including KDD, ICDM, and SDM. He received the Best Application Paper Award on KDD-2008, the Best Research Paper Award on ICDM-2011, and the Best of SDM-2015. His research is supported by the US National Science Foundation for Distinguished Young Scholars of China.
\end{IEEEbiography}

\end{document}